%% file: main.tex
\pdfoutput=1

\documentclass[11pt]{article}

\usepackage[]{ex-acl}

\usepackage{times}
\usepackage{latexsym}

\usepackage[T1]{fontenc}
\usepackage[utf8]{inputenc}
\usepackage{microtype}
\usepackage{inconsolata}

\input{imp-def.tex}

\title{Assessing Logical Puzzle Solving in Large Language Models:\\Insights from a Minesweeper Case Study}

\author{
  Yinghao Li,\quad Haorui Wang,\quad Chao Zhang\\
  Georgia Institute of Technology, Atlanta, USA\\
  \texttt{$\{$yinghaoli, hwang984, chaozhang$\}$@gatech.edu}
}

\begin{document}
\maketitle

\input{sections/0.abs.tex}

\input{sections/1.intro.tex}

\input{sections/2.related.works.tex}

\input{sections/3.setup.tex}

\input{sections/4.results.tex}

\input{sections/5.conclusion.tex}

\input{sections/6.limitations.tex}

\input{sections/9.ack.tex}

\bibliography{reference}

\clearpage

\appendix

\input{appendix/1.setup.tex}

\input{appendix/2.results.tex}

\end{document}

%% file: imp-def.tex
\usepackage{times}
\usepackage{latexsym}
\usepackage{cite}
\usepackage{natbib}

\usepackage[T1]{fontenc}

\usepackage{microtype}

\usepackage{amsmath}
\usepackage{amsfonts}
\usepackage{graphicx}  
\usepackage{xspace}
\usepackage{eufrak}
\usepackage{bm}
\usepackage{multirow}
\usepackage{booktabs}
\usepackage{subfig}
\usepackage[utf8]{inputenc}

\usepackage{cleveref}
\usepackage{enumitem}
\usepackage{adjustbox}

\usepackage{xargs}
\usepackage{etoolbox}
\usepackage[flushleft]{threeparttable}

\usepackage[colorinlistoftodos,prependcaption,textsize=tiny]{todonotes}
\usepackage[group-separator={,}, group-minimum-digits=4]{siunitx}
\usepackage{comment}

\usepackage{listings}
\usepackage{xcolor}

\definecolor{codegreen}{rgb}{0,0.6,0}
\definecolor{codegray}{rgb}{0.5,0.5,0.5}
\definecolor{codepurple}{rgb}{0.58,0,0.82}
\definecolor{backcolour}{rgb}{0.95,0.95,0.95}

\lstdefinestyle{mystyle}{
    backgroundcolor=\color{backcolour},   
    commentstyle=\color{codegreen},
    keywordstyle=\color{magenta},
    numberstyle=\tiny\color{codegray},
    stringstyle=\color{codepurple},
    basicstyle=\ttfamily\scriptsize,
    breakatwhitespace=false,         
    breaklines=true,                 
    captionpos=t,                    
    keepspaces=true,                 
    numbers=left,                    
    numbersep=5pt,                  
    showspaces=false,                
    showstringspaces=false,
    showtabs=false,                  
    tabsize=2
}

\setlist{nosep}

\crefformat{section}{\S~#2#1#3} 
\crefformat{subsection}{\S~#2#1#3}
\crefformat{subsubsection}{\S~#2#1#3}

\newcommand{\safedefine}[2]{
  \ifdef{#1}{\renewcommand{#1}{#2}}{\newcommand{#1}{#2}}
}
\makeatletter
\newcommand\footnoteref[1]{\protected@xdef\@thefnmark{\ref{#1}}\@footnotemark}
\makeatother

\newcommand*{\rom}[1]{\uppercase\expandafter{\romannumeral #1}}

\newcommandx{\chao}[2][1=]{\todo[linecolor=red,backgroundcolor=red!25,bordercolor=red,#1]{#2}}
\newcommandx{\yli}[2][1=]{\todo[linecolor=blue,backgroundcolor=blue!25,bordercolor=blue,#1]{#2}}
\newcommandx{\thiswillnotshow}[2][1=]{\todo[disable,#1]{#2}}

\newcommand{\ie}{\emph{i.e.}\xspace} 
\newcommand{\eg}{\emph{e.g.}\xspace} 

\safedefine{\iff}{\emph{i.f.f.}\xspace}
\safedefine{\iid}{\emph{i.i.d.}\xspace}

\newcommand{\flagged}{\texttt{F}}
\newcommand{\blank}{\texttt{.}}
\newcommand{\numbered}[1]{\texttt{#1}}
\newcommand{\unopened}{\texttt{?}}
\newcommand{\lquote}{\texttt{\textasciigrave}}
\newcommand{\rquote}{\texttt{\textquotesingle}}

%% file: sections/0.abs.tex
\begin{abstract}


Large Language Models (LLMs) have shown remarkable proficiency in language understanding and have been successfully applied to a variety of real-world tasks through task-specific fine-tuning or prompt engineering.
Despite these advancements, it remains an open question whether LLMs are fundamentally capable of reasoning and planning, or if they primarily rely on recalling and synthesizing information from their training data.
In our research, we introduce a novel task---Minesweeper---specifically designed in a format unfamiliar to LLMs and absent from their training datasets.
This task challenges LLMs to identify the locations of mines based on numerical clues provided by adjacent opened cells.
Successfully completing this task requires an understanding of each cell's state, discerning spatial relationships between the clues and mines, and strategizing actions based on logical deductions drawn from the arrangement of the cells.
Our experiments, including trials with the advanced GPT-4 model, indicate that while LLMs possess the foundational abilities required for this task, they struggle to integrate these into a coherent, multi-step logical reasoning process needed to solve Minesweeper.
These findings highlight the need for further research to understand the nature of reasoning capabilities in LLMs under similar circumstances, and to explore pathways towards more sophisticated AI reasoning and planning models.

\end{abstract}

%% file: sections/1.intro.tex
\section{Introduction}
\label{sec:intro}


Large Language Models (LLMs) have made remarkable strides in the Natural Language Processing (NLP) arena, capturing the spotlight with their multifaceted capabilities.
These models have been effectively utilized across a spectrum of NLP tasks, including information extraction \citep{Agarwal.2022.LLM.Clinical.IE, Zhu.2023.LLM.IR.Survey}, answering arithmetic and common-sense questions \citep{Li.2022.systematic.investigation, Yuan.2023.arithmetic.tasks, Imani.2023.MathPrompter}, as well as aiding in strategic planning \citep{Yao.2023.ReAct, Wang.2023.describe.explain.plan.select} or acting as game agents \citep{Callison-Burch.2022.dungeons.and.dragons, Wang.2023.voyager, Gupta.2023.chatgpt.poker.players}.
With growing scale, LLMs begin to exhibit ``emergent abilities'' \citep{Wei.2022.emergent.abilities, Shaeffer.2023.emergent.abilities}, including marked improvements in their ability to follow instructions, perform multi-step reasoning and planning, and even extend to comprehending humor \citep{OpenAI.2023.GPT4}.
Such flourishing capability has triggered excitement within the research community, cultivating the belief that LLMs could be instrumental in achieving Artificial General Intelligence \citep{Sebastian.2023.sparks.of.artificial.general.intelligence, Zhang.2023.chatgpt.survey.aigc, Tu.2023.towards.generalist.biomedical.ai}.
Nevertheless, a crucial question remains underexplored:
\emph{To what extent can the reasoning capabilities of LLMs go beyond the scope of their training distribution?}

This question is critical as it uncovers if LLMs are reliable for sophisticated hands-off planning in various scenarios such as biomedical experimental design \citep{Bram.2023.ChemCrow, ODonoghue.2023.BioPlanner} and autonomous driving \citep{Sha.2023.LanguageMPC}, where the agents have to deal with off-distribution inputs occasionally.
Unfortunately, the answer is not straightforward, and the debate is going on \citep{Huang.2022.self.improve, Helbling.2023.LLM.self.defense, Wu.2023.SPRING, Huang.2023.LLM.self.correct, Valmeekam.2023.self.critiquing.plans, Stechly.2023.iterative.prompting}.
The reason is that LLMs' substantial parameter counts enable them to store and recall vast amounts of information from their training material \citep{Gudibande.2023.the.false.promise, Rosenfeld.2020.a.constructive.prediction, Kaplan.2020.scaling.laws, Brown.2020.language.models}, which may contain conventional reasoning datasets such as GSM8K \citep{Cobbe.2021.gsm8k}, MultiArith \citep{Roy.2015.MultiArith}, or StrategyQA \citep{Geva.2021.StrategyQA}.
This could exaggerate LLMs' reasoning abilities on such benchmarks \citep{Qin.2023.chatgpt.general.purpose, Deng.2023.data.contamination}, leading to false a promise when applying LLMs to practical scenarios.

Although attempts to develop new benchmarks that challenge LLMs suggest potential zero-shot learning successes \citep{Suzgun.2023.challenging.big.bench, Bao.2023.OOD.logical.reasoning}, there is an underlying concern that these benchmarks may not be distinct enough from the LLMs' training data, thereby giving LLMs an undue advantage and masking their true abilities \citep{Valmeekam.2022.LLMs.cant.plan}.
For instance, \citet{Gudibande.2023.the.false.promise} suggest that smaller LLMs only learn specific tasks when fine-tuned with data from larger models and that these improvements do not generalize well.
Drawing an analogy between smaller and larger LLMs to humans implies that LLMs may not exhibit genuine intelligence or intrinsic reasoning beyond their training distribution \citep{Liu.2023.evaluating.logical.reasoning, Sebastian.2023.sparks.of.artificial.general.intelligence}.
To robustly claim that LLMs possess reasoning abilities, they must be tested under a wider array of conditions.
A fundamental approach to overcoming these challenges is to decouple the test inputs from the LLMs' pre-training data, compelling them to rely on task descriptions for reasoning and planning rather than prior knowledge.
This entails an environment with clear instructions, a rich but well-defined action space, and a distinct objective.
Such a setup facilitates the verification of the reasoning process and qualitative performance assessment, while it also reduces the chances of successful random guessing.

Minesweeper, a well-known logic puzzle game illustrated in Figure~\ref{fig:s2-board}, is proposed as an ideal testing ground under these considerations.
The game presents a range of complexities: while its fundamental principles are straightforward and accessible to beginners, achieving proficiency requires advanced logical reasoning skills that go beyond mere pattern recognition on the game board.
For LLMs, the challenge in Minesweeper lies in interpreting the states of individual cells, making use of numerical hints, and understanding the spatial interconnections between cells.
This is essential for accurately determining the locations of mines and strategizing subsequent moves based on incomplete information—--a process that is instinctive for human players.
Despite LLMs' prior knowledge of the universally known rules of Minesweeper, the transformation of game boards from visual representations to machine-readable text has been limited, with no existing format aligning with our representation.
Consequently, LLMs must rely on their intrinsic logic for problem-solving by understanding and following the rules, as pattern matching from training data is impractical.

\input{sections/figures/f1.board.tex}

Building upon our initial motivation, we conduct a comprehensive set of experiments with LLMs including GPT-3.5-16k, GPT-3.5-instruct, and GPT-4 \citep{OpenAI.2022.chatgpt, OpenAI.2023.GPT4}, probing their inherent reasoning skills.
We introduce the Minesweeper game in various formats to determine how different types of input influence LLM performance.
Objective measures like the count and proportion of accurate moves, correctly flagged mines, and completely resolved boards were employed to gauge effectiveness.
However, a critical aspect of our evaluation is the manual examination of the LLMs' reasoning processes, inspecting the validity and logical soundness of their intermediate deductions.
Our findings indicate that these models generally struggle to maintain consistent logical reasoning chains.
The GPT-3.5 versions, in particular, tended to repeat information from provided examples or previous conversations, often failing to adjust to the updated board layouts.
While GPT-4 showed improvements in response diversity, relevance, and coherence, it still faced issues with logical inconsistencies and incomplete reasoning.
Our research categorizes errors in a detailed simulated environment, which sheds light on the specific skills required for solving logical puzzles, underscoring the importance of these competencies in future LLM-focused studies.
Moreover, we notice that LLMs tend to generate ad-hoc sequences of actions rather than following logical reasoning chains based on action history, a finding not commonly reported, providing novel insights into LLM operational behaviors.
To support forthcoming research and comprehensive evaluation of LLMs, we have published our code and data at \href{https://github.com/Yinghao-Li/Minesweeper-for-LLM}{https://github.com/Yinghao-Li/Minesweeper-for-LLM}.
We hope it contributes to LLMs' broader understanding and future development.

%% file: sections/figures/f1.board.tex
\begin{figure*}[tbp]
  \centering{
    \subfloat[]{
      \label{subfig:minesweeper.gui}
      \includegraphics[height=1.8 in]{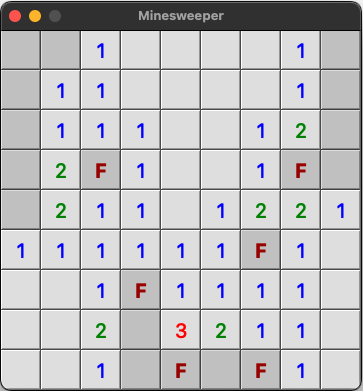}
    }
    \hfill
    \subfloat[]{
      \label{subfig:minesweeper.table}
      \includegraphics[height=1.8 in]{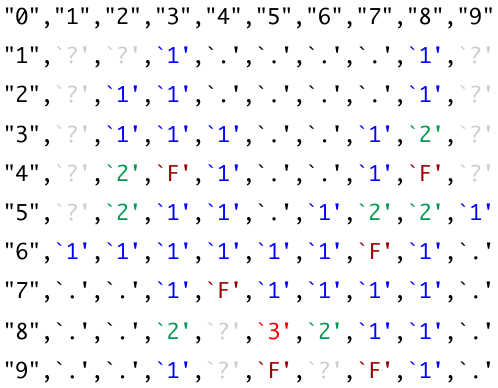}
    }
    \hfill
    \subfloat[]{
      \label{subfig:minesweeper.coord}
      \includegraphics[height=1.8in]{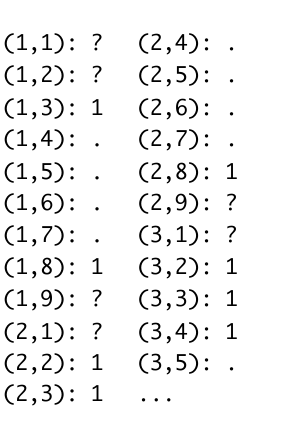}
    }
    \hfill
    \subfloat[]{
      \label{subfig:minesweeper.action.history}
      \includegraphics[height=1.8in]{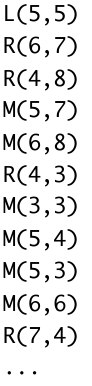}
    }
  }
  \caption{
    An example of Minesweeper on a $9\times9$ board containing \num{10} mines, along with its interaction format.
    Subfigure~\ref{subfig:minesweeper.gui} displays the game's GUI;
    Subfigure~\ref{subfig:minesweeper.table} shows a plain-text, table-formatted representation of the game board, enhanced with color for improved visualization;
    Subfigure~\ref{subfig:minesweeper.coord} depicts the coordinate-based plain-text representation of the board; and
    Subfigure~\ref{subfig:minesweeper.action.history} provides a log of the player's (in this case, the first author's) actions, where ``\texttt{L}'', ``\texttt{R}'', and ``\texttt{M}'' denote left-click, right-click, and middle-click actions, respectively.
  }
  \label{fig:s2-board}
\end{figure*}

%% file: sections/2.related.works.tex
\section{Related Works}
\label{sec:related.works}

Recent studies have explored the reasoning capabilities of LLMs, with several investigations underway to understand \citep{Qin.2023.chatgpt.general.purpose, Bang.2023.chatgpt.evaluation, Liu.2023.evaluating.logical.reasoning, Xu.2023.logical.reasoning, Srivastava.2023.beyond.imitation.game, Bao.2023.OOD.logical.reasoning}, enhance \citep{Wei.2022.COT, Imani.2023.MathPrompter}, and leverage \citep{Yao.2023.ReAct, Sha.2023.LanguageMPC} these abilities for various applications.
Some critics argue that the datasets used for evaluation are overly simplistic and potentially compromised by inclusion in the training data of LLMs \citep{OpenAI.2023.GPT4, Wei.2023.Skywork}, casting doubt on the authenticity of these purported capabilities \citep{Valmeekam.2022.LLMs.cant.plan, Bao.2023.OOD.logical.reasoning}.
In response, new tasks and evaluation frameworks have been suggested \citep{Valmeekam.2022.LLMs.cant.plan, Xu.2023.logical.reasoning, Bao.2023.OOD.logical.reasoning, Vervoort.2023.hypothetic.deductive.reasoning, Wang.2023.voyager}.
However, these tend to focus narrowly on specific aspects of arithmetic or commonsense reasoning, often presented in unstructured natural language that intersects with LLMs' pre-existing world knowledge.
This overlap presents a challenge in fully assessing LLMs' inductive reasoning abilities.
In contrast, the game of Minesweeper encompasses symbolic, arithmetic, and commonsense reasoning in a structured format lying outside the scope of LLMs' training data, and thus demands genuine reasoning skills for successful resolution.

Alone another track, the use of LLMs in gameplay has gained attention, with efforts to harness them for diverse games such as card games \citep{Gupta.2023.chatgpt.poker.players, Guo.2023.suspicion.agent}, interactive narrative games \citep{Callison-Burch.2022.dungeons.and.dragons, Tsai.2023.large.language.models.text.games, Cui.2023.thespian, Zhu.2023.CALYPSO}, chess \citep{Noever.2020.ChessTransformer, Stockl.2021.watching, Suzgun.2023.challenging.big.bench, Feng.2023.ChessGPT}, and video games like Minecraft \citep{Wang.2023.voyager, Wang.2023.DEPS}.
Nonetheless, we contend that such games are intrinsically distinct from Minesweeper in both structure and the level of challenge they pose to LLMs.
Card and interactive narrative games largely draw on common sense and narrative understanding, with a constrained range of actions.
This limitation can obscure whether LLM decisions stem from deep reasoning or instinctual response.
Chess, while offering a more extensive range of possibilities, presents an abstract playfield that typically requires tailored pre-training for LLMs to grasp its rules.
Minecraft poses a further challenge with its complex syntax and the need for meticulous, often biased, prompt engineering, making it an impressive display but not an accurate indicator of LLMs' functional application.
Minesweeper, by comparison, offers a straightforward yet strategically rich gameplay experience with clear objectives and direct evaluation criteria, making it an exemplary model for assessing the practical reasoning capabilities of LLMs.

%% file: sections/3.setup.tex
\section{Minesweeper}
\label{sec:minesweeper}

Minesweeper, a classic logic game, has been a part of the Windows operating system since around 1990s.
Though the rule is simple, it features a rich and discrete action space, a clear goal, and direct feedbacks and evaluation metrics.
The objective is straightforward: players must unveil all cells on a board peppered with hidden mines without detonating any.
Each cell on the board can exist in one of four possible states (Figure~\ref{subfig:minesweeper.table}):

\begin{itemize}[leftmargin=*]
    \item \textbf{Unopened} (``\unopened''): The state of these cells is unknown; they could transition into any of the other three states.
    \item \textbf{Numbered} (``\numbered{1}'' to ``\numbered{8}''): These cells reveal the count of adjacent mines, including diagonals, providing critical clues for safe navigation.
    \item \textbf{Blank} (``\blank''): These cells have no adjacent mines, often opening up larger areas of the board.
    \item \textbf{Flagged} (``\flagged''): Cells suspected to contain mines are marked accordingly for safety.
\end{itemize}
Notably, the choice of symbols is arbitrary and should not impact the model performance, as they serve as mere representations of concepts.

During gameplay, players may execute one of three actions per turn: left-click (``\texttt{L}'') to reveal a cell's contents, right-click (``\texttt{R}'') to flag a suspected mine, and middle-click (``\texttt{M}'') to verify the correctness of flagged cells based on adjacent numbers.
On a beginner's $9\times9$ board (Figure~\ref{subfig:minesweeper.gui}), this results in \num{243} potential actions per round (Figure~\ref{subfig:minesweeper.action.history}), creating an extensive action landscape despite numerous actions being invalid (as listed in Table~\ref{tb3.feedback}).
The game provides immediate feedback for each action, with outcomes ranging from revealing or flagging cells to error messages, game endings or victories, allowing models to adjust their strategies in real-time.
Victory is reached by correctly identifying and flagging all mines or revealing all non-mine cells, whereas failure occurs upon missteps such as erroneous left-click on mines or incorrect flag placements during middle-click verifications.
With action space and game objective established, metrics such as the proportion of valid moves, the percentage of successfully solved games, and the average number of moves to resolve a board can be readily computed.
In summary, Minesweeper serves as a less knowledge-intensive, more symbol-comprehension and basic math-reasoning benchmark, pivoting on spatial reasoning over pure information retrieval.

We employ two plain-text formats to present the game board when interacting with LLMs, aiming to understand LLMs' capabilities of comprehending each of them and to reduce the potential influence of input format on the reasoning process.
The first format is \textbf{table representation}, depicted in Figure~\ref{subfig:minesweeper.table}.
Here, the game board is portrayed as a table with the states of each cell enclosed in \LaTeX-style quotation marks (grave accent ``\lquote'' and apostrophe ``\rquote'') to differentiate them from the cell indices and separators.
Rows and columns are delineated by line break ``\texttt{\textbackslash n}'' and comma ``\texttt{,}'' respectively, with indices for the first row and column serving as the coordinates for the cells.
This format is intuitive for human interpretation;
LLMs equipped with capabilities to understand tables, as suggested by the literature \citep{Chen.2023.large.language.model, Singha.2023.tabular.representation}, should process it similarly.
For comparison, we also employ \textbf{coordinate representation} (Figure~\ref{subfig:minesweeper.coord}).
It explicitly associates each cell's coordinates with its state in the form of a look-up table, potentially offering a more direct description for LLMs to comprehend the board's configuration.

The following sections first explore two critical skills required for Minesweeper agents (\cref{sec:board.understanding}).
Then, they evaluate the proficiency of LLMs in playing Minesweeper, detailing objective scores and case studies that highlight their reasoning and planning capabilities (\cref{subsec:gameplay.results}).
Each section begins with a brief introduction of the experimental setup, with comprehensive details available in \cref{appsec:experiment.setup}.

%% file: sections/4.results.tex
\input{sections/tables/tb1.board.understanding.tex}

\section{Board Understanding}
\label{sec:board.understanding}

\subsection{Experiment Setup}
A fundamental skill for LLMs in Minesweeper is the ability to comprehend the game board.
To assess this capability, we conducted two types of straightforward experiments: \textbf{board navigation} and \textbf{neighbor counting}.
In board navigation, LLMs are presented with a game board and a specific coordinate, and they are tasked to identify the corresponding cell state.
For neighbor counting, LLMs are given an additional arbitrary state from the set $\{$\lquote\unopened\rquote, \lquote\blank\rquote, \lquote\numbered{1}\rquote, \lquote\numbered{2}\rquote,$\}$ and are required to calculate the number of occurrences of this state (ranging from \num{0} to \num{8}) surrounding the provided coordinate.
We annotated \num{100} randomly generated $9\times9$ boards, each containing \num{10} mines, with complete action steps (refer to Figure~\ref{subfig:minesweeper.action.history}).
The boards were randomized at different stages of the game, either fully or partially revealed, and \num{3} sets of coordinates were selected randomly for the experiments.

\subsection{Results}
The performance of GPT-3.5-16k and GPT-3.5-instruct was evaluated using both table and coordinate representations, as shown in Table~\ref{tb1.board.understanding}.
With the table representation, GPT-3.5-16k demonstrated a notable deficiency in accurately identifying the cell state, failing approximately one-third of the time in board navigation tasks.
GPT-3.5-instruct showed even lower performance.
Attempts to enhance performance through various modifications to the boards and prompts, including removing indices, adding examples, and requesting models to revise their outputs, were unsuccessful.
However, with the coordinate representation, both models exhibited improved results.
This suggests that the challenge of interpreting tables might be a limiting factor for the models in subsequent gameplay tests.
Nonetheless, GPT-3.5 displayed a failure rate of over $10\%$, which is suboptimal given that the task essentially involves copying and pasting values in coordinate-state mappings.
In neighbor counting tasks, both models achieved a maximum accuracy of about one-third, significantly better than random guessing.
This indicates that GPT-3.5 models possess basic arithmetic and geographical planning skills, which are essential for Minesweeper.
These findings suggest that the models have a moderate level of board understanding, which is sufficient to not severely impede the gameplay tests that follow.

\input{sections/tables/tb2.game.performance.tex}

\section{Minesweeper Gameplay}
\label{subsec:gameplay.results}

\subsection{Experiment Setup}
In our study, we utilize $5\times5$ Minesweeper boards containing \num{4} hidden mines, a setup less complex than the conventional beginner level.
LLMs are instructed to start the game by left-clicking the center cell and then proceed with either left, right, or middle-click actions on cells they assess as beneficial for advancing gameplay.
We craft and annotate \num{100} boards for GPT-3.5 models, ensuring the initial left-click would reveal at least \num{10} cells ($40\%$ of the board).
This strategy aims to streamline the length of the prompts and minimize the number of interactive rounds required. 
For GPT-4, we randomly subsample \num{10} boards from the pool to reduce the experiment time and expense.

Prompting techniques such as chain-of-thought reasoning \citep{Wei.2022.COT} and few-shot in-context learning (\cref{subsec:game.playing}) are employed to activate the LLMs' reasoning capabilities and provide sufficient information for decision-making.
Each game session permits a maximum of \num{10} actions per board, concluding either when the model achieves victory, triggers a mine, or produces unrecognizable responses.
Given that human annotators, on average, complete each board with an average of $6.14$ actions, we deemed a $10$-action limit adequate for the model to effectively resolve most boards, contingent on its reasoning ability.

\input{sections/figures/f3.prompting.modes.tex}

To balance the influence of prompt details and accommodate models with shorter prompt constraints, our study introduces two distinct prompting approaches: 1) \textbf{Natural Conversation} (NC) mode, and 2) \textbf{Compact History} (CH) mode, as illustrated in Figure \ref{fig:s4.prompting.modes}.
In the NC mode, each game commences with a detailed explanation of rules and examples.
During each round, we prompt LLMs to execute actions along with their associated reasoning.
The game system then provides feedback on these actions and updates the board in a new user message, while maintaining the complete history of the interaction.
CH mode, on the other hand, condenses all relevant information and interaction history into a concise, singular prompt, eliminating the need for extensive conversational details.
This approach significantly shortens the length of the conversation history.
Given the turn-based nature of Minesweeper, the CH mode is effective in providing all vital information required for the model to make well-informed decisions.

\subsection{Metrics}
The evaluation of our model's performance is anchored in objective metrics, which encompass: the count of valid actions, the accurate identification of mines, the number of successfully completed games, and the logical coherence of reasoning chains.
A ``valid'' action is one that advances gameplay without activating a mine.
Table~\ref{tb3.feedback} details a broad spectrum of negative examples.
The metric for correctly flagged mines is based on the quantity of mines identified by the model that align with the actual mine locations at the game's conclusion.
A game is deemed ``solved'' when all mines are accurately marked and no extraneous cells are flagged.
To assess the reasoning chains, we initially select \num{5} game boards where each model executed the highest number of valid actions.
We then perform a manual inspection to ensure these chains are logically sound and directly pertinent to the gameplay.
Due to the variable action counts across different models and boards, the totals of actions and reasoning chains differ, as Table~\ref{tb2.game.performance} illustrates.

\input{sections/figures/f4.instr.actions.tex}

\subsection{Results and Discussion}

Table~\ref{tb2.game.performance} showcases the performance comparisons between various GPT family models and a human benchmark, focusing on valid actions, correctly identified mines, completed games, and logical reasoning chains.
In line with the findings on board comprehension, models utilizing coordinate representation exhibit superior performance across all metrics compared to table representation.
This supports our previous assertion that LLMs struggle with tables containing symbols not found in their training data, preferring familiar in-domain terminology as evidenced in the studies by \citet{Chen.2023.large.language.model} and \citet{Singha.2023.tabular.representation}.

\input{sections/figures/f2.case.study.tex}

\paragraph{Objective Scores}
In assessing specific gameplay metrics, we first focus on three main areas: Game Outcomes, Mine Identification, and Action Accuracy.
Among the GPT-3.5 variants, 3.5-instruct demonstrates better overall performance compared to 3.5-16k in these metrics, albeit with a marginally increased failure rate, which diverges from the board understanding results.
A detailed analysis of GPT-3.5-instruct's \emph{initial} move reveals a consistent anomaly: the model persistently selects the \texttt{R(2,2)} action, contrary to the given instruction of \texttt{L(3,3)} (refer to Table~\ref{tb3.feedback} for invalid actions and Listing~\ref{ls:init.game.prompt}, Line 83 for the specific prompt).
After receiving feedback, it switches to the \texttt{L(1,1)} action.
This action often uncovers fewer cells in our specially selected boards, usually just the chosen cell itself (as illustrated in Figure~\ref{fig:s4.instr.actions}), which increases the probability of the next action being valid.
This behavior could also explain the higher failure rate and lower repeatance rate observed in GPT-3.5-instruct compared to the 3.5-16k model.
Furthermore, GPT-3.5-instruct tends to make conservative moves by flagging all cells on the board, which accounts for its inability to solve any board despite having a higher ratio of flagged mines.
Overall, all GPT-3.5 models present a noticeable pattern of repetitive actions during board solving, persisting even with direct instructions to avoid repetition (see Listing~\ref{ls:init.game.prompt}, line 29).
This repetitive behavior suggests a significant limitation in the models' capacity to fully understand and reason about the nuanced variations occurring on the board at each turn.
As Transformer-based models designed primarily for next-token prediction \citep{Vaswani.2017.transformer}, their focus likely remains on static aspects of cell states, overlooking subtle changes.
This attentional bias, compounded by the lack of pre-training on tasks resembling this one, often leads to the repeated execution of identical actions.

In contrast, GPT-4 exhibits a marked improvement in following instructions, achieving a zero rate of repetition while maintaining a high frequency of valid actions.
Notably, it flags the most mines, despite not solving any board and recording a high percentage of game failures, which is partly attributed to its tendency for divergent actions.
When both operated in NC mode, GPT-4 is more prone to generating unrecognizable actions compared to GPT-3.5-16k, as illustrated in Figure~\ref{fig:s4.gpt4.reasoning}, even when the conversation history is within the token limit.
A plausible interpretation of this behavior is that GPT-4 demonstrates heightened sensitivity to recent changes in board states, allocating more attention to these immediate alterations.
Conversely, it appears to pay less regard to the more distant historical context, where the constraints governing action formats are established.

\input{sections/figures/f5.gpt4.reasoning.tex}

\paragraph{Reasoning Chains}
The foregoing discussion highlights that merely counting valid actions is not an adequate measure of the models' reasoning skills.
Therefore, we shift our focus to the logical coherence of their reasoning chains.
A review of the final section in Table~\ref{tb2.game.performance} reveals that the GPT-3.5 variants struggle to generate coherent reasoning.
For instance, in a typical case from GPT-3.5-16k, shown in Figure~\ref{fig:s4.case.study}, the model correctly identifies cell \lquote\numbered{4}\rquote at \texttt{(2,4)} and understands that there are four mines around, but fails to grasp the concept of ``neighboring cells'' or to accurately count the surrounding cell states, as indicated by the statement marked in red.
Furthermore, the statement in golden font is not well-grounded, making it challenging to discern any logical connection between its first and second halves.
This sentence appears more like an unsuccessful attempt to replicate our in-context example ``\emph{Given that there's just one adjacent mine, it's logical to deduce that the unopened cell at (2,3) contains the mine}'' (refer to Listing~\ref{ls:init.game.prompt}, Line 51) rather than a display of genuine reasoning.
While this example is specifically for GPT-3.5-16k, similar errors are prevalent across all GPT-3.5 outputs.
It appears that despite possessing a certain degree of individual capabilities in elementary symbolic understanding, geographical reasoning, and arithmetic calculation, as suggested by the board understanding results, the GPT-3.5 models fall short in integrating these skills cohesively for action planning, resulting in a fragmented reasoning chain.

GPT-4 exhibits more encouraging results compared to its predecessors.
While its achievements are modest, GPT-4 demonstrates a certain level of logical understanding in this unfamiliar task, suggesting an ability to grasp the game's rules and dynamics beyond mere replication.
To explore the extent and origin of this reasoning, we delve deeper into GPT-4's reasoning chains, analyzing a representative case in Figure~\ref{fig:s4.gpt4.reasoning}.
It shows that GPT-4 can logically reason if the number of neighboring unopened cells equals the remaining unflagged mines near a numbered cell, specifically in simple scenarios like \lquote\numbered{1}\rquote and \lquote\numbered{2}\rquote, and particularly around the board's corners (\eg, \texttt{(1,2)} or \texttt{(5,4)}).
Yet, even in such basic situations, GPT-4 shows around $50\%$ fails.
The model struggles particularly with configurations where exactly three unopened cells surround a \lquote\numbered{3}\rquote cell and other complex setups, as seen in the lower rows of the example.
To determine if GPT-4's ability stems from the rules or the examples, we conducted a control experiment by omitting Example 1 from the initial prompt (Listing~\ref{ls:init.game.prompt}, Line 43), which closely mirrors the successful scenarios and is the sole example of left-clicking.
This led to a dramatic $90\%$ reduction in coherent reasoning with the coordinate board representation, indicating the significant role of examples in GPT-4's reasoning process, likely by providing a familiar context for reference and imitation.

Another notable point from Figure~\ref{fig:s4.gpt4.reasoning} is GPT-4's non-linear action sequence.
For a human player, following the initial action of \texttt{R(1,2)}, the logical next step would be to mark the adjacent unopened cell as a mine, \ie, \texttt{R(1,1)}.
However, GPT-4 instead opts for \texttt{R(5,1)}, an irrational choice that strays from the focus of the previous action, before returning to \texttt{(1,1)}.
This behavior suggests that GPT-4 might not fully consider its historical actions and lacks the capability for long-term, multi-step planning, a crucial aspect in evaluating an agent's intrinsic reasoning abilities as opposed to mere summarization capabilities.

It is important to note that the results discussed were derived from experiments conducted on $5\times5$ boards, and it is reasonable to anticipate that performance may further deteriorate on larger boards with more flexible mine arrangements.

%% file: sections/tables/tb1.board.understanding.tex
\begin{table}[t]\small
  \centering
  \begin{tabular}{clcc}
    \toprule
    Task & Representation & 3.5-16k & 3.5-instr \\
    \midrule
    \multirow[c]{6}{*}{\shortstack{Board\\Navigation}} & table & 66.7 & 53.3 \\
      & \ $-$ ids & 63.7 & 40.7 \\
      & \ $+$ example & 64.3 & 47.0 \\
      & \ $+$ verification & 66.3 & - \\
      \cmidrule{2-4}
      & coordinate & 82.3 & 66.7 \\
      & \ $+$ example & \textbf{89.7} & 61.0 \\
    \midrule
    \midrule
    \multirow[c]{2}{*}{\shortstack{Neighbor\\Counting}} 
      & table & 33.3 & 16.7\\
      & coordinate & \textbf{37.3} & 30.3 \\
    \bottomrule
  \end{tabular}
  \caption{
    Comparative analysis of GPT-3.5 variants in board understanding tasks.
    The performance are quantified using exact-matching accuracy percentages.
    The notation ``$-$ ids'' indicates the omission of indices in the table representation, while ``$+$ example'' denotes the inclusion of additional examples in the prompts.
    The ``$+$ verification'' symbol refers to the application of the self-verification technique as described in \citep{Weng.2022.self.verification}.
    As GPT-3.5-instruct is not optimized for conversational contexts, the self-verification is not applicable.
  }
  \label{tb1.board.understanding}
\end{table}

%% file: sections/tables/tb2.game.performance.tex
\begin{table*}[t]\small
  \centering
  \begin{tabular}{l|c|cc|cc|cc|cc}
    \toprule
    \multirow[c]{2}{*}{} & Human & \multicolumn{2}{c|}{3.5-16k (NC)} & \multicolumn{2}{c|}{3.5-16k (CH)} & \multicolumn{2}{c|}{3.5-instr (CH)} & \multicolumn{2}{c}{4 (NC)}\\
    \cmidrule(lr){2-10}
    & GUI & Table & Coord & Table & Coord & Table & Coord & Table & Coord\\
    \midrule
    \multicolumn{10}{c}{\textbf{Game Outcome}} \\
    \midrule
    \# Total Games & \multicolumn{7}{c|}{100} & \multicolumn{2}{c}{10} \\
    \midrule
    $\bullet$ \% Solved ($\uparrow$)& 89.0  & 0.0 & 0.0  & 0.0  & 0.0  & 0.0  & 0.0 & 0.0 & 0.0 \\
    $\bullet$ \% Failed ($\downarrow$) & 11.0 & 17.0  & 35.0  & 17.0  & 35.0  & 19.0  & 41.0 & 70.0 & 70.0 \\
    \midrule
    \multicolumn{10}{c}{\textbf{Mine Identification}} \\
    \midrule
    \# Total Mines & \multicolumn{7}{c|}{400} & \multicolumn{2}{c}{40} \\
    \midrule
    $\bullet$ \% Flagged ($\uparrow$)& 93.0 & 2.8 & 11.7 & 8.0 & 9.5 & 11.7 & 23.3 & 30.0 & 45.0 \\
    \midrule
    \multicolumn{10}{c}{\textbf{Action Accuracy}} \\
    \midrule
    \# Total Actions& 514 & 825 & 679 & 700 & 687 & 813 & 642 & 41 & 52 \\
    \midrule
    $\bullet$ \% Valid ($\uparrow$)  & 99.6 & 7.2 & 22.4 & 15.7 & 22.6 & 24.1 & 64.6 & 82.9 & 82.7 \\
    $\bullet$ \% Repeated ($\downarrow$)& 5.1 & 43.6 & 44.6 & 42.3 & 45.3 & 39.4 & 19.9 & 0.0 & 0.0 \\
    \midrule
    \multicolumn{10}{c}{\textbf{Reasoning Soundness}} \\
    \midrule
    \# Selected Reasoning Chains & - & 34 & 38 & 45 & 39 & 45 & 45 & 25 & 38 \\
    \midrule
    $\bullet$ \% Valid Actions ($\uparrow$) & - & 61.8 & 81.1 & 46.7 & 59.0 & 80.0 & 95.6 & 88.0 & 84.2 \\
    $\bullet$ \% Accurate and Coherent Logic ($\uparrow$) & - & 0.0 & 0.0  & 0.0 & 0.0 & 0.0 & 0.0 & 12.0 & 26.4 \\
    \bottomrule
  \end{tabular}
  \caption{
    The comparison of GPT model variants, prompting modes, and board representation formats on $5\times5$ boards with $4$ mines.
    ``\#'' indicates the number of instances; ``\%'' represents the percentage ratio.
    The initial action, \texttt{L(3,3)}, is consistently excluded from the count.
    GPT-3.5-instruct is prompted using the compact history mode, as it is not optimized for chat-based interactions.
    The terms ``Table'' and ``Coord'' in the table refer to the table-format and coordinate-format representations of the Minesweeper board, respectively.
    The percentages of solved and failed cases do not sum to 100\% due to instances where the game was not completed within \num{10} steps.
    The repetition is calculated independently for each board.
  }
  \label{tb2.game.performance}
\end{table*}

%% file: sections/figures/f3.prompting.modes.tex
\begin{figure}[t!]
  \centering
  \includegraphics[width=0.96\columnwidth]{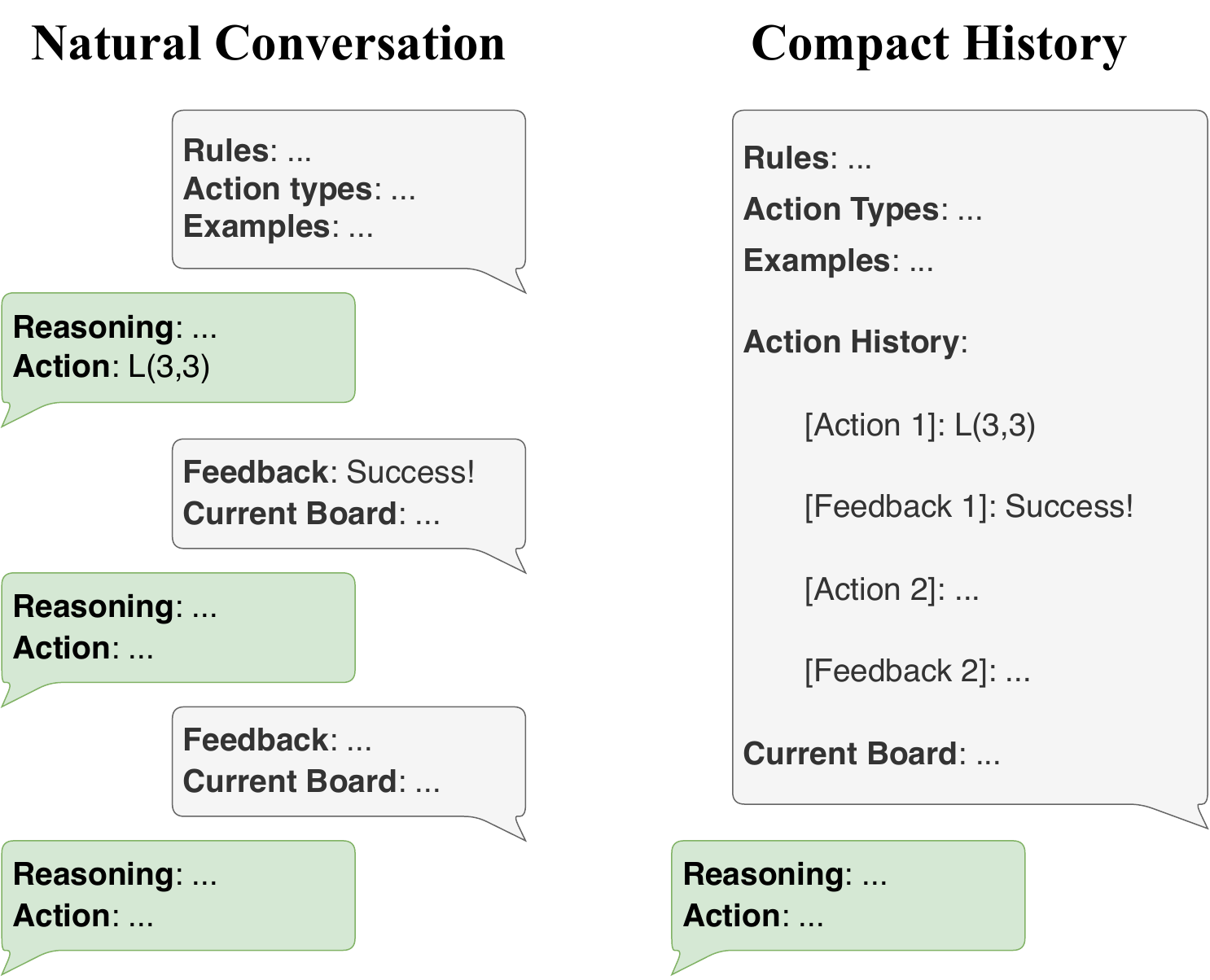}
  \caption{
    Interaction prompting modes.
    The ``Natural Conversation'' mode encompasses the full interaction history, whereas the ``Compact History'' mode condenses the actions generated by the LLM and the game's feedback into a succinct, unified prompt.
  }
  \label{fig:s4.prompting.modes}
\end{figure}

%% file: sections/figures/f4.instr.actions.tex
\begin{figure*}[t!]
    \centering
    \includegraphics[width=\textwidth]{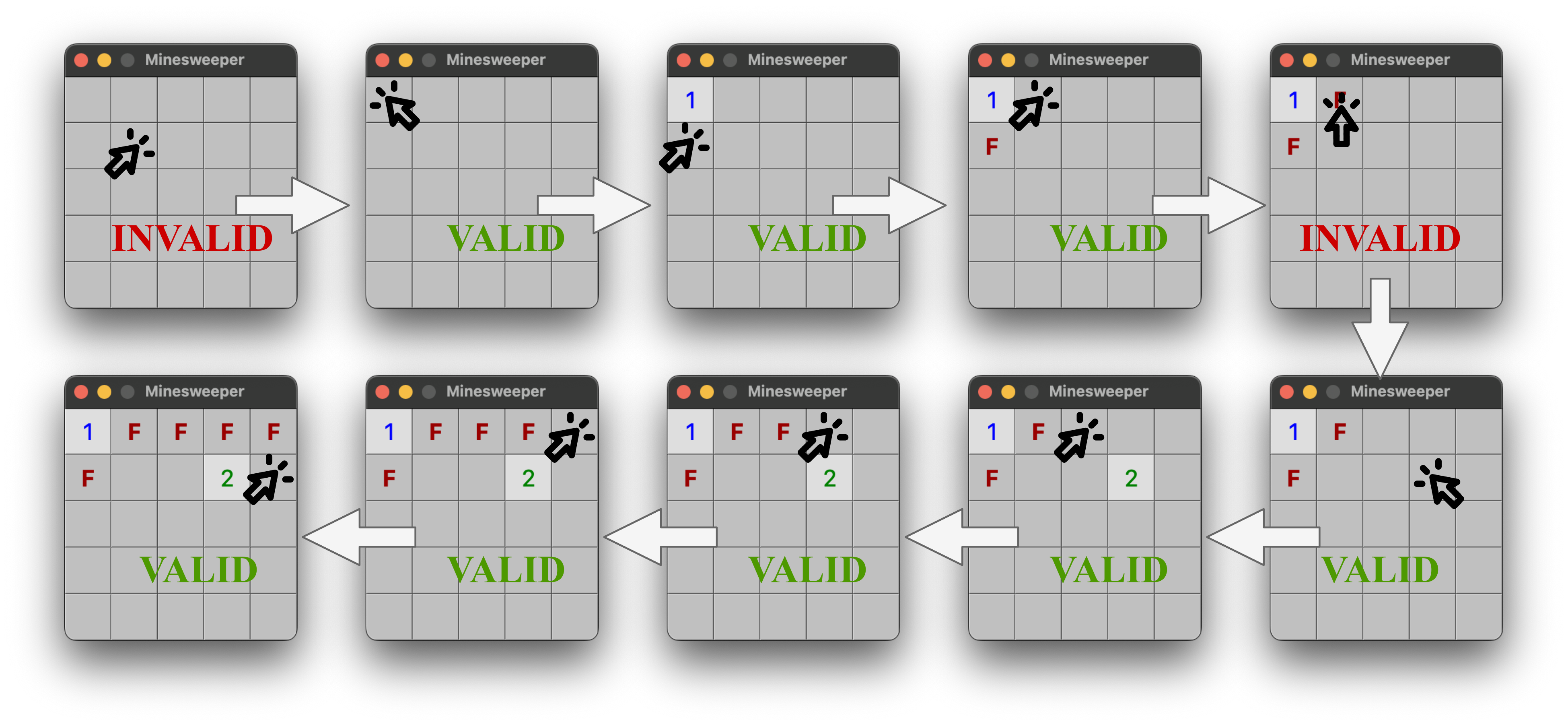}
    \caption{
      A detailed analysis of the example interactions performed by GPT-3.5-instruct.
      Arrows oriented to the left and right signify left and right mouse clicks, respectively.
      The arrow pointing upwards represents a middle-click.
      The majority actions are technically allowed but do not effectively advance the gameplay.
    }
    \label{fig:s4.instr.actions}
  \end{figure*}

%% file: sections/figures/f2.case.study.tex
\begin{figure}[t!]
    \centering
    \includegraphics[width=\columnwidth]{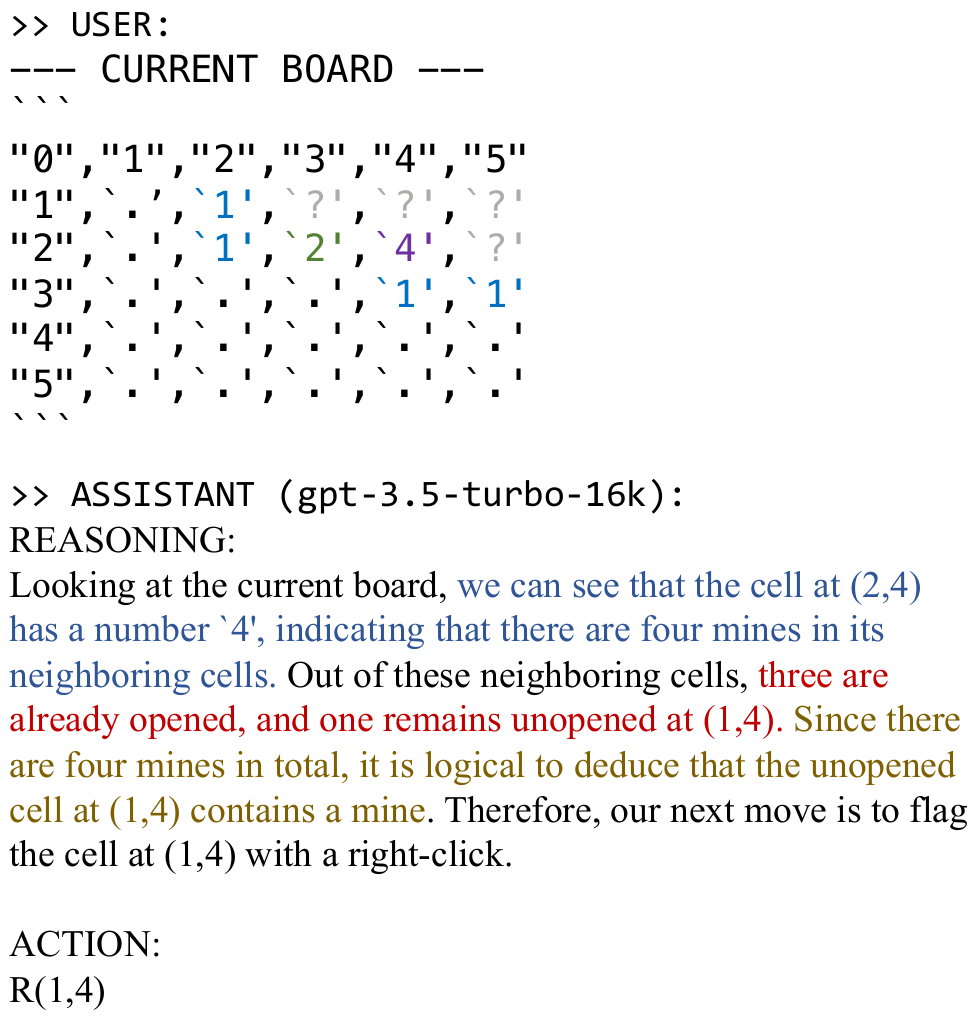}
    \caption{
      A case study of a ``valid'' action and its corresponding reasoning generated by GPT-3.5-16k for solving Minesweeper.
      Blue indicates logical reasoning; red and golden are illogical ones. 
    }
    \label{fig:s4.case.study}
  \end{figure}

%% file: sections/figures/f5.gpt4.reasoning.tex
\begin{figure*}[t!]
    \centering
    \includegraphics[width=\textwidth]{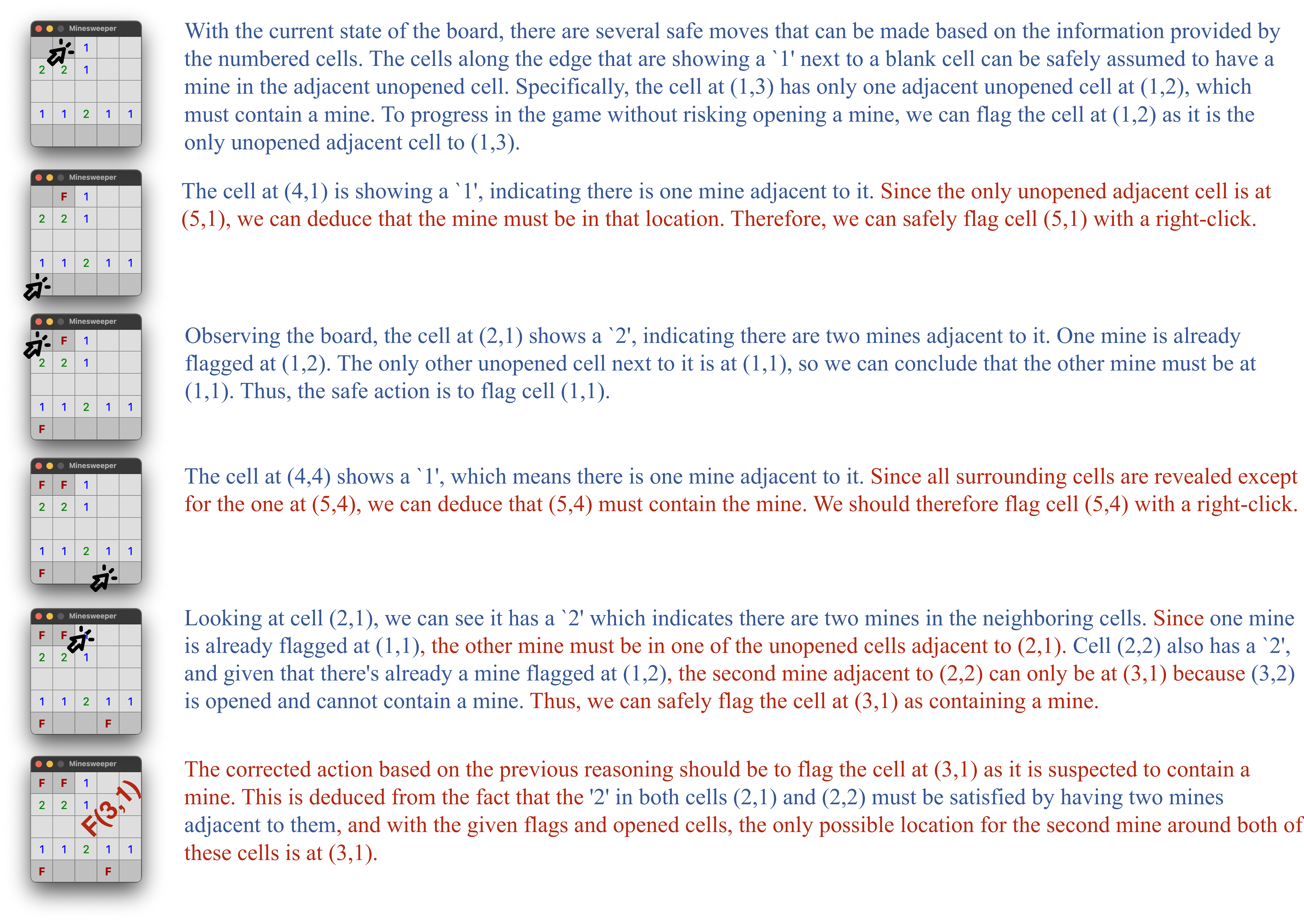}
    \caption{
      This figure presents a case study of the reasoning sequences formulated by GPT-4 with coordinate board representation during action planning.
      The GUI on the left shows the board states and actions taken by the agent during the game.
      Elements highlighted in blue represent accurate facts and logical inferences as assessed by human evaluation, while those marked in red indicate incorrect observations or illogical conclusions.
      Notably, the final generated action ``\texttt{F(3,1)}'' deviates from the permissible action formats, resulting in the termination of the game.
    }
    \label{fig:s4.gpt4.reasoning}
  \end{figure*}

%% file: sections/5.conclusion.tex
\section{Conclusion}
\label{sec:conclusion}

In this study, we assessed the capabilities of LLMs in solving logical puzzles by examining their performance in the game Minesweeper, a task for which they have received limited exposure during the pre-training phase.
This approach aims to assess the intrinsic reasoning capabilities of LLMs.
Our experiments reveal that GPT-3.5 models exhibit basic abilities in spatial navigation, symbol recognition, and counting.
However, they struggle with multi-step planning and generating coherent reasoning chains.
GPT-4 shows improvement in simpler scenarios with shorter logic chains but still encounters issues with hallucinations and inconsistent context awareness.
We find that GPT-4 can learn from examples and apply this knowledge to similar situations but is less adept at deducing underlying rules and applying them to novel scenarios, unlike GPT-3.5 models which tend to replicate examples.
Our conclusion seems to be in correlation with \citet{Yadlowsky.2023.pretraining.data.mixtures}:
LLMs do not produce new knowledge (a situation where logical reasoning is necessary); they are just (good) handlers of existing experience.

We do not intend to diminish the significance of LLMs or their transformative impact on natural language processing.
Rather, we believe it is premature to consider LLMs as an intelligent object and a threat to human society.
They resemble a comprehensive dictionary, useful for reference but lacking in comprehension.
We hope that our work can inspire future research on this topic, and we will continue exploring the capabilities of LLMs.

%% file: sections/6.limitations.tex
\section*{Limitations}

In our study, we employed two board representation methods: table and coordinate representations.
Our results indicate that coordinate representation is more effective for LLM comprehension and offers a wider range for evaluating the model's reasoning abilities.
However, we did not explore other board representation methods, such as different table formats (\eg, HTML, Markdown) or image-based approaches.
According to \citet{Singha.2023.tabular.representation}, alternative table formats may not significantly differ in effectiveness from our chosen method.
Still, it's possible that other representations could be more suitable for our specific context.
Moreover, our initial attempts to use an image-based Web interface with board GUI were not successful.
Therefore, we focus on text-based representations in this study.
While advanced prompting methods like Program of Thoughts (PoT, \citealp{Chen.2023.PoT}) or Tree of Thoughts (ToT, \citealp{Yao.2023.ToT}) may boost performance, their use could divert from assessing the true reasoning abilities of LLMs and their applicability in simplifying complex issues without extensive external input.
Hence, we deferred the application of these sophisticated techniques to future studies.

Another area for exploration is fine-tuning LLMs like Llame-2-70b \citep{Touvron.2023.Llama2} with specific reasoning capabilities on our dataset labeled with human actions.
The idea is that if a fine-tuned model outperforms the original or GPT-4 when the fine-tuning and testing data match in terms of symbol usage, board size, and representation, and does not improve with variations in these factors, we can infer that the ``reasoning capability'' is tied to the training data and not broadly generalizable.
However, due to constraints in computational resources, we haven't conducted these experiments, leaving them for future research.

At last, we would like to emphasize that we can easily extend the Minesweeper dataset to a scale of $n_{\rm row}\times n_{\rm column} \choose n_{\rm mine}$ with zero effort.
Nonetheless, we keep a relatively small scale of test cases as it is sufficient to support our conclusion. 
With the future development of LLMs, a larger evaluation set might be desired for a more robust and unbiased objective metric-based evaluation.

%% file: sections/9.ack.tex
\section*{Acknowledgments}

This work was supported in part by NSF IIS-2008334, IIS-2144338, and ONR MURI N00014-17-1-2656.

%% file: appendix/1.setup.tex
\section{Experiment Setup}
\label{appsec:experiment.setup}

\subsection{GPT Versions}


In this paper, we discuss three distinct versions of the GPT model: GPT-3.5-16k, GPT-3.5-instruct, and GPT-4.
The term GPT-3.5-16k is used to denote the ``gpt-3.5-turbo-16k'' model as listed on the OpenAI API model webpage, specifically the checkpointed 0613.\footnote{\href{https://platform.openai.com/docs/models}{https://platform.openai.com/docs/models}}
The GPT-3.5-instruct model refers to the ``gpt-3.5-turbo-instruct'' variant, with the checkpoint version dated 2023-09-15.
Lastly, GPT-4 represents the ``gpt-4'' model, utilizing also the 0613 checkpoint version.
All mentioned checkpoints are hosted on Microsoft Azure,\footnote{\ *.openai.azure.com}.
The model temperature is set to \num{0} in all cases.

\subsection{Board Understanding}
\label{subsec:board.understanding}

In our board understanding experiments, we initially generate \num{1000} randomly arranged boards, each featuring a $9\times9$ grid with \num{10} hidden mines.
Notably, the central cells, positioned at \texttt{(5,5)}, are always mine-free.
From this pool, we select \num{100} boards, ensuring that the initial click at the center \texttt{L(5,5)} reveals at least \num{10} cells.
Subsequently, we conduct gameplay using the graphical user interface depicted in Figure~\ref{subfig:minesweeper.gui}.
The actions of the annotators, who are average players in Minesweeper and unaware of the board's layout beyond the initial instruction to click the center, are recorded as shown in Figure~\ref{subfig:minesweeper.action.history}.
Although not all trials are successful—--with about $90\%$ boards being solved and the remainder failing due to human errors or 50-50 guesses—--the entire action history, including unsuccessful attempts, is preserved for analysis.

For each board, we randomly choose an action (excluding the first and last from the history) and associate board state as the focus.
We then select \num{3} random coordinates per board, setting two challenges for LLMs: to determine the status of each selected cell (board navigation), and to count the number of specific symbols within $\{$\lquote\unopened\rquote, \lquote\blank\rquote, \lquote\numbered{1}\rquote, \lquote\numbered{2}\rquote$\}$ in their neighbors, including diagonally adjacent ones (neighbor counting).
This process yields \num{300} test cases for each task.

As outlined in \cref{sec:board.understanding}, we employ two distinct board representations: table and coordinate.
Each required a slightly different approach in prompting.
Specifically, the table representation uses \LaTeX-style quotation marks, while the coordinate representation does not use quotations for board cell states and employs regular double quotes in the necessary positions in the game description.
Examples illustrating these differences are provided below.

\begin{lstlisting}[caption=An example of board navigation prompting with table representation.]
You will be presented with a 9 by 9 board for the Minesweeper game. The board is wrapped in a 10 by 10 table, where the first row and the first column with numbers in double quotation marks are the row and column indices. A coordinate (x,y) represents the cell at the x-th row and y-th column, where x and y, starting from 1, are the row and column indices, respectively. The state of each cell is represented by the following symbols:
- `.' represents a blank cell.
- `1' to `8' represents numbered cells with that number of mines in the adjacent cells.
- `F' represents a flagged cell.
- `?' represents an unopened cell.

--- EXAMPLES ---
--- PARTIAL BOARD ---
"0","1","2","3","4"
"1",`?',`?',`1',`.'
"2",`?',`?',`3',`1'
"3",`?',`F',`2',`F'
"4",`?',`2',`2',`1'

QUESTION: What is the cell at coordinate (1,3)?
ANSWER: `1'

--- PARTIAL BOARD ---
"0","1","2","3","4"
"1",`?',`?',`1',`.'
"2",`?',`?',`3',`1'
"3",`?',`F',`2',`F'
"4",`?',`2',`2',`1'

QUESTION: What is the cell at coordinate (4,1)?
ANSWER: `?'

--- END OF EXAMPLES ---

--- CURRENT BOARD ---
"0","1","2","3","4","5","6","7","8","9"
"1",`?',`?',`?',`?',`1',`.',`.',`.',`.'
"2",`?',`?',`?',`?',`1',`.',`1',`1',`1'
"3",`?',`1',`1',`1',`1',`.',`1',`F',`1'
"4",`?',`1',`.',`.',`.',`.',`1',`1',`1'
"5",`?',`2',`1',`1',`.',`.',`.',`.',`.'
"6",`?',`?',`?',`1',`1',`2',`3',`2',`1'
"7",`?',`?',`?',`?',`?',`?',`?',`?',`?'
"8",`?',`?',`?',`?',`?',`?',`?',`?',`?'
"9",`?',`?',`?',`?',`?',`?',`?',`?',`?'

QUESTION: What is the cell at coordinate (2,1)?
ANSWER: 
\end{lstlisting}
\begin{lstlisting}[caption=An example of board navigation prompting with coordinate representation.]
You will be presented with a 9 by 9 board for the Minesweeper game, which is devided into cells. The cells are presented as "coordinate: state" mappings. A coordinate (x,y) represents the element at the x-th row and y-th column, where x and y, starting from 1, are the row and column indices, respectively. The state of each cell is represented by the following symbols:
- "." represents a blank cell.
- "1" to "8" represents numbered cells with that number of mines in the adjacent cells.
- "F" represents a flagged cell.
- "?" represents an unopened cell.

--- EXAMPLES ---
--- PARTIAL BOARD ---
(1,1): ?
(1,2): ?
(1,3): ?
(1,4): ?
(2,1): ?
(2,2): ?
(2,3): ?
(2,4): ?
(3,1): ?
(3,2): ?
(3,3): ?
(3,4): 1
(4,1): ?
(4,2): ?
(4,3): ?
(4,4): 1

QUESTION: What is the cell at coordinate (3,4)?
ANSWER: "1""

--- END OF EXAMPLES ---

--- CURRENT BOARD ---
(1,1): ?
(1,2): ?
(1,3): ?
(1,4): ?
(1,5): 1
(1,6): .
(1,7): .
(1,8): .
(1,9): .
(2,1): ?
(2,2): ?
(2,3): ?
(2,4): ?
(2,5): 1
(2,6): .
(2,7): 1
(2,8): 1
(2,9): 1
(3,1): ?
(3,2): 1
(3,3): 1
(3,4): 1
(3,5): 1
(3,6): .
(3,7): 1
(3,8): F
(3,9): 1
(4,1): ?
(4,2): 1
(4,3): .
(4,4): .
(4,5): .
(4,6): .
(4,7): 1
(4,8): 1
(4,9): 1
(5,1): ?
(5,2): 2
(5,3): 1
(5,4): 1
(5,5): .
(5,6): .
(5,7): .
(5,8): .
(5,9): .
(6,1): ?
(6,2): ?
(6,3): ?
(6,4): 1
(6,5): 1
(6,6): 2
(6,7): 3
(6,8): 2
(6,9): 1
(7,1): ?
(7,2): ?
(7,3): ?
(7,4): ?
(7,5): ?
(7,6): ?
(7,7): ?
(7,8): ?
(7,9): ?
(8,1): ?
(8,2): ?
(8,3): ?
(8,4): ?
(8,5): ?
(8,6): ?
(8,7): ?
(8,8): ?
(8,9): ?
(9,1): ?
(9,2): ?
(9,3): ?
(9,4): ?
(9,5): ?
(9,6): ?
(9,7): ?
(9,8): ?
(9,9): ?

QUESTION: What is the cell at coordinate (2,1)?
ANSWER: 
\end{lstlisting}
\begin{lstlisting}[caption=An example of neighbor counting prompting with table representation., label=ls:nc.table]
You will be presented with a 9 by 9 board for the Minesweeper game. The board is wrapped in a 10 by 10 table, where the first row and the first column with numbers in double quotation marks are the row and column indices. A coordinate (x,y) represents the cell at the x-th row and y-th column, where x and y, starting from 1, are the row and column indices, respectively. The state of each cell is represented by the following symbols:
- `.' represents a blank cell.
- `1' to `8' represents numbered cells with that number of mines in the adjacent cells.
- `F' represents a flagged cell.
- `?' represents an unopened cell.

--- EXAMPLES ---
--- PARTIAL BOARD ---
"0","1","2","3","4","5"
"1",`?',`?',`?',`?',`?'
"2",`?',`?',`?',`?',`?'
"3",`?',`?',`?',`1',`1'
"4",`?',`?',`?',`1',`.'
"5",`?',`?',`?',`1',`.'

QUESTION: How many cells `F' are neighbors (including diagonal) of the cell with coordinate (2,1)?
ANSWER:

To find out how many cells with the value `1' are neighbors of the cell with coordinate (2,1), we need to look at the 8 neighboring cells of (2,1). These coordinates are:
(1,1), (1,2), (3,1), (3,2), (2,2).

Now, we will check the values of these cells on the given Minesweeper board:

(1,1) = `?'
(1,2) = `?'
(3,1) = `?'
(3,2) = `?'
(2,2) = `?'

All of these neighboring cells are `?'. So, the number of cells with `1' that are neighbors of the cell (2,1) is:

ANSWER: 0.

--- END OF EXAMPLES ---

--- CURRENT BOARD ---
"0","1","2","3","4","5","6","7","8","9"
"1",`?',`?',`?',`?',`1',`.',`.',`.',`.'
"2",`?',`?',`?',`?',`1',`.',`1',`1',`1'
"3",`?',`1',`1',`1',`1',`.',`1',`F',`1'
"4",`?',`1',`.',`.',`.',`.',`1',`1',`1'
"5",`?',`2',`1',`1',`.',`.',`.',`.',`.'
"6",`?',`?',`?',`1',`1',`2',`3',`2',`1'
"7",`?',`?',`?',`?',`?',`?',`?',`?',`?'
"8",`?',`?',`?',`?',`?',`?',`?',`?',`?'
"9",`?',`?',`?',`?',`?',`?',`?',`?',`?'

QUESTION: How many cells `1' are neighbors (including diagonal) of the cell with coordinate (2,1)?
Let's think step by step.
ANSWER: 
\end{lstlisting}
\begin{lstlisting}[caption=An example of neighbor counting prompting with coordinate representation., label=ls:nc.coord]
You will be presented with a 9 by 9 board for the Minesweeper game, which is devided into cells. The cells are presented as "coordinate: state" mappings. A coordinate (x,y) represents the element at the x-th row and y-th column, where x and y, starting from 1, are the row and column indices, respectively. The state of each cell is represented by the following symbols:
- "." represents a blank cell.
- "1" to "8" represents numbered cells with that number of mines in the adjacent cells.
- "F" represents a flagged cell.
- "?" represents an unopened cell.

--- EXAMPLES ---
--- PARTIAL BOARD ---
(1,1): F
(1,2): ?
(1,3): ?
(1,4): F
(2,1): ?
(2,2): ?
(2,3): ?
(2,4): ?
(3,1): ?
(3,2): F
(3,3): ?
(3,4): 1
(4,1): ?
(4,2): ?
(4,3): ?
(4,4): 1

QUESTION: How many cells "F" are neighboring (including diagonally) the cell with coordinates (2,1)?
Let's think step by step
ANSWER:

To find out how many cells with the value "1" are neighbors of the cell with coordinate (2,1), we need to look at the 8 neighboring cells of (2,1). These coordinates are:
(1,1), (1,2), (3,1), (3,2), (2,2).

Now, we will check the values of these cells on the given Minesweeper board:

(1,1) = F
(1,2) = ?
(3,1) = ?
(3,2) = F
(2,2) = ?

From of these neighboring cells (1,1) and (3,2) are "F". So, the number of cells with "1" that are neighbors of the cell (2,1) is:

ANSWER: 2.

--- END OF EXAMPLES ---

--- CURRENT BOARD ---
(1,1): ?
(1,2): ?
(1,3): ?
(1,4): ?
(1,5): 1
(1,6): .
(1,7): .
(1,8): .
(1,9): .
(2,1): ?
(2,2): ?
(2,3): ?
(2,4): ?
(2,5): 1
(2,6): .
(2,7): 1
(2,8): 1
(2,9): 1
(3,1): ?
(3,2): 1
(3,3): 1
(3,4): 1
(3,5): 1
(3,6): .
(3,7): 1
(3,8): F
(3,9): 1
(4,1): ?
(4,2): 1
(4,3): .
(4,4): .
(4,5): .
(4,6): .
(4,7): 1
(4,8): 1
(4,9): 1
(5,1): ?
(5,2): 2
(5,3): 1
(5,4): 1
(5,5): .
(5,6): .
(5,7): .
(5,8): .
(5,9): .
(6,1): ?
(6,2): ?
(6,3): ?
(6,4): 1
(6,5): 1
(6,6): 2
(6,7): 3
(6,8): 2
(6,9): 1
(7,1): ?
(7,2): ?
(7,3): ?
(7,4): ?
(7,5): ?
(7,6): ?
(7,7): ?
(7,8): ?
(7,9): ?
(8,1): ?
(8,2): ?
(8,3): ?
(8,4): ?
(8,5): ?
(8,6): ?
(8,7): ?
(8,8): ?
(8,9): ?
(9,1): ?
(9,2): ?
(9,3): ?
(9,4): ?
(9,5): ?
(9,6): ?
(9,7): ?
(9,8): ?
(9,9): ?

QUESTION: How many cells "1" are neighbors (including diagonal) of the cell with coordinate (2,1)?
Let's think step by step.
ANSWER: 
\end{lstlisting}

In our analysis of the prompts, we employ the Chain-of-Thought (COT) technique, as described by \citet{Wei.2022.COT}, for the neighbor counting task but not for board navigation.
The rationale behind this is that neighbor counting necessitates multi-step action: initially identifying the neighboring cells of a given target cell and subsequently tallying a specific symbol within these neighbors.
In contrast, the board navigation task is straightforward, involving a direct query about a specified target cell.
We have observed that incorporating COT into the neighbor counting process notably enhances its performance.

Furthermore, we have made the inclusion of in-context examples optional for the board navigation task, but mandatory for neighbor counting.
The need for examples in neighbor counting arises from the models' propensity to generate unclear outputs in this task.
By providing an example, we aim to guide the models towards producing more interpretable responses.

\input{appendix/tables/tb3.invalid.actions.tex}
\subsection{Minesweeper}
\label{subsec:game.playing}


The detailed setup of the Minesweeper experiment is presented in \cref{subsec:gameplay.results}.
We hereby provide the actual initial prompts with table-formatted board representation used in our experiments.
The coordinate representation closely mirrors the variations found in Listing~\ref{ls:nc.table} and Listing~\ref{ls:nc.coord} and will not be omitted.

\begin{lstlisting}[caption=Initial prompt for Minesweeper gameplay., label=ls:init.game.prompt]
>> SYSTEM:
You are a helpful assistant who is good at playing Minesweeper.

>> USER:
--- MINESWEEPER INTRODUCTION ---
In Minesweeper, 4 hidden mines are scattered throughout a 5 by 5 board, which is divided into cells. The rows are seperated by newlines, and columns by commas. The board is structured as a 6 by 6 table, with the first row and column labeled using numbers in double quotation marks to indicate row and column indices. Cells have multiple possible states:
- Unopened cells (represented by `?', which cover the board at the start of the game, can also be made by removing flags)
- Numbered cells (represented by `1' to `8', which indicate the number of mines in the eight neighboring cells, including those diagonally adjacent)
- Blank cells (represented by `.', which have no neighboring mines)
- Flagged cells (represented by `F', which are marked by the player to indicate a potential mine location)

A player selects a cell to open it. If a player opens a cell containing a mine, the game ends in a loss. Otherwise, the opened cell displays either a number, indicating the number of mines diagonally and/or adjacent to it, or a blank tile (sometimes shown as a 0), and all adjacent cells will automatically be opened. To win a game of Minesweeper, all non-mine cells must be opened without opening a mine.

--- ACTION OPTIONS ---
There are three permissible actions in Minesweeper:

- Left-click an unopened cell (`?') to reveal it.
- Right-click an unopened cell (`?') to place a flag or a flagged cell (`F') to remove the flag.
- Middle-click on a numbered cell (`1' to `8') to unveil its neighboring cells, but only if all adjacent mines have been correctly flagged. If any flags are misplaced, you'll lose the game.

--- ACTION FORMAT ---
Each of your actions should be formatted as "A(row,col)", where:
- "A" represents the action type: "L" denotes a left-click, "R" indicates a right-click, and "M" signifies a middle-click.
- "row" specifies the row number of the targeted cell.
- "col" details the column number of the targeted cell.
For instance, an action like "L(1,2)" translates to a left-click on the cell located at the first row and second column.

please ensure:
- You do not duplicate actions.
- You submit only one action at a time.

--- CURRENT BOARD ---
```
"0","1","2","3","4","5"
"1",`?',`?',`?',`?',`?'
"2",`?',`?',`?',`?',`?'
"3",`?',`?',`?',`?',`?'
"4",`?',`?',`?',`?',`?'
"5",`?',`?',`?',`?',`?'
```

--- EXAMPLES ---
Example 1:
--- PARTIAL BOARD ---
"0","1","2","3","4"
"1",`.',`1',`?',`?'
"2",`.',`1',`?',`?'
"3",`.',`1',`2',`4'
"4",`.',`.',`.',`2'

REASONING: The cell located at (3,2) displays the number `1', indicating there's a single mine in its neighboring cells. Among these neighbors, 7 are opened as empty or numbered, and only one remains unopened at (2,3). Given that there's just one adjacent mine, it's logical to deduce that the unopened cell at (2,3) contains the mine. Thus, our next move is to flag the cell at (2,3) with a right-click.

ACTION: R(2,3)

Example 2:
--- PARTIAL BOARD ---
"0","4","5","6","7"
"1",`.',`1',`?',`?'
"2",`.',`1',`F',`?'
"3",`.',`1',`2',`4'
"4",`.',`.',`.',`2'

REASONING: The cell at (1,5) displays the number `1', indicating a single adjacent mine. Upon examining its neighboring cells, only the cell at (2,6) is flagged with `F'. This suggests that all mines neighboring (1,5) have been identified. Consequently, we can safely unveil the other surrounding cells by middle-clicking on (1,5).

ACTION: M(1,5)

Example 3:
--- PARTIAL BOARD ---
"0","1","2","3","4"
"1",`1',`1',`2',`?'
"2",`?',`?',`?',`?'
"3",`?',`?',`?',`?'

REASONING: The cell at (1,1) indicates there's a single mine amongst its neighbors. Examining the cells adjacent to it, both (2,1) and (2,2) remain unopened, implying one of them contains a mine. Similarly, the cell at (1,2) displays a `1', suggesting that out of (2,1), (2,2), and (2,3), one holds a mine. Since one of (2,1) or (2,2) already contains a mine, it becomes evident that (2,3) is mine-free. We can then safely uncover (2,3) with a left-click.

ACTION: L(2,3)

--- END OF EXAMPLES ---

--- RESPONSE GUIDE ---
Let's think step by step.

Begin by detailing your rationale in the "REASONING" section, followed by specifying your move using the previously mentioned format in the "ACTION" section. Ensure your reasoning doesn't exceed 200 words. To commence the game, left-click the center cell at (3, 3).

REASONING:
ACTION:
\end{lstlisting}


In subsequent iterations, the prompts are modified based on the model's responses.
When the model generates a valid action, the updated game board is presented, followed by a request for the model to initiate the next move.
In contrast, if the action is invalid, an error message is displayed, prompting the model to select an alternative action.
Table~\ref{tb3.feedback} enumerates the various types of invalid actions and their corresponding feedback tailored for LLMs.
Note that these actions do not directly violate the game rules.
Rather, they are deemed invalid within the scope of our experiments because they are logically ungrounded and do not contribute to the progression of the game.

%% file: appendix/tables/tb3.invalid.actions.tex
\begin{table*}[t]\small
  \centering
  \begin{tabular}{p{0.22\linewidth}|p{0.2\linewidth}|p{0.5\linewidth}}
    \toprule
    Type of Invalid Actions & Example & Feedback \\
    \midrule
    \multicolumn{3}{c}{Coordinate} \\
    \midrule
    Coordinate Out of Bound & ``\texttt{L(0,10)}'' on a $9\times 9$ board & Invalid Coordinates! Please make sure your coordinate are within [1, 9] for rows and [1, 9] for columns. \\
    \midrule
    \multicolumn{3}{c}{Game Initialization} \\
    \midrule
    Starting by Right-Clicking & ``\texttt{R(5,5)}'' on a new board & Please begin by left-clicking on the center cell. \\
    Starting by Middle-Clicking & ``\texttt{M(5,5)}'' on a new board & Please begin by left-clicking on the center cell. \\
    \midrule
    \multicolumn{3}{c}{Left-Clicking} \\
    \midrule
    Left-Clicking on Blank Cell & ``\texttt{L(5,5)}'' where (5,5) is opened as \texttt{\textasciigrave.\textquotesingle} & Invalid action: Cannot left-click a blank cell. Left-click is only for unopened cells (\texttt{\textasciigrave?\textquotesingle}). \\
    \midrule
    Left-Clicking on Flagged Cell & ``\texttt{L(5,5)}'' where (5,5) is flagged as \texttt{\textasciigrave F\textquotesingle} & Invalid action: Cannot left-click a flagged cell. Left-click is only for unopened cells (\texttt{\textasciigrave?\textquotesingle}). \\
    \midrule
    Left-Clicking on Numbered Cell & ``\texttt{L(5,5)}'' where (5,5) is opened as \texttt{\textasciigrave 2\textquotesingle} & Invalid action: Cannot left-click a numbered cell. Left-click is only for unopened cells (\texttt{\textasciigrave?\textquotesingle}). \\
    \midrule
    \multicolumn{3}{c}{Middle-Clicking} \\
    \midrule
    Middle-Clicking on Blank Cell & ``\texttt{M(5,5)}'' where (5,5) is opened as \texttt{\textasciigrave.\textquotesingle} & Invalid action: Cannot middle-click a blank cell. Middle-click is only for numbered cells (\texttt{\textasciigrave 1\textquotesingle} to \texttt{\textasciigrave 8\textquotesingle}). \\
    \midrule
    Middle-Clicking on Flagged Cell & ``\texttt{M(5,5)}'' where (5,5) is flagged as \texttt{\textasciigrave F\textquotesingle} & Invalid action: Cannot middle-click a flagged cell. Middle-click is only for numbered cells (\texttt{\textasciigrave 1\textquotesingle} to \texttt{\textasciigrave 8\textquotesingle}). \\
    \midrule
    Middle-Clicking on Unopened Cell & ``\texttt{M(5,5)}'' where (5,5) is unopened (\texttt{\textasciigrave?\textquotesingle}) & Invalid action: Cannot middle-click an unopened cell. Middle-click is only for numbered cells (\texttt{\textasciigrave 1\textquotesingle} to \texttt{\textasciigrave 8\textquotesingle}). \\
    \midrule
    Middle-Clicking When Numbered Cell Has no Flagged Neighbor &  & Error: No flagged cells detected nearby. Flag adjacent mines before middle-clicking. \\
    \midrule
    Middle-Clicking When \# Flagged Neighbor Mismatchs Cell Number &  & Error: Flag count mismatch. Ensure all adjacent mines are flagged before middle-clicking. \\
    \midrule
    \multicolumn{3}{c}{Right-Clicking} \\
    \midrule
    Right-Clicking on Blank Cell & ``\texttt{R(5,5)}'' where (5,5) is opened as \texttt{\textasciigrave.\textquotesingle} & Invalid action: Cannot right-click a blank cell. Right-click is only for unopened cells (\texttt{\textasciigrave?\textquotesingle}) or flagged cells (\texttt{\textasciigrave?\textquotesingle}). \\
    \midrule
    Right-Clicking on Numbered Cell & ``\texttt{R(5,5)}'' where (5,5) is opened as \texttt{\textasciigrave 1\textquotesingle} & Invalid action: Cannot right-click a numbered cell. Right-click is only for unopened cells (\texttt{\textasciigrave?\textquotesingle}) or flagged cells (\texttt{\textasciigrave?\textquotesingle}). \\
    \bottomrule
  \end{tabular}
  \caption{
    Types and feedbacks for invalid actions.
  }
  \label{tb3.feedback}
\end{table*}

%% file: appendix/2.results.tex
\input{appendix/tables/tb4.ablation.tex}

\section{Obfuscation and Numerical Expression}

To study the impact of existing Minesweeper-related description within LLMs' training materials, we conduct additional experiments to assess the impact of varying the game descriptions and the representation of numerical values on the performance of the model.
In the obfuscation experiment, we replaced all instances of ``Minesweeper'' with descriptions of a generic logical puzzle, substituting ``mine'' with ``thorn'' and standard mouse-click actions within game description and action feedbacks with abstract symbols (\eg, ``L/R/M'').
In the other set of experiments, Arabic numerals that indicate the numbers of surrounding mines (\eg, \numbered{1},\numbered{2},$\dots$,\numbered{8}) were converted to their Roman counterparts (\eg, \numbered{I}, \numbered{II}, $\cdots$, \numbered{VIII}), with other elements remaining unchanged.
The outcomes are summarized in Table~\ref{tb:ablation-study}.
They show that these variations have minimal effect on model performance, indicating that these factors do not significantly hinder the model's reasoning abilities.
Our findings suggest that the general exposure to Minesweeper game rules across the internet does not provide a distinct advantage for GPT models in solving these puzzles, and our assumption of ``Minesweeper being a new challenge for GPTs'' holds.

\section{Interaction Examples}
\label{appsec:results.and.analysis}

In this section, we provide more case studies on the reasoning chains generated by LLMs as support of our discussion in \cref{subsec:gameplay.results}.
\textbf{For better readability, we convert all coordinate representations to table representations.}
Please refer to our \href{https://github.com/Yinghao-Li/Minesweeper-for-LLM}{GitHub repository} for a complete set of experiment results.

\begin{lstlisting}[caption=An example of LLM-system interaction with table representation in NC mode from GPT-3.5-16k., label=ls:gpt3.5.game.results]
// Initial Instructions are Omitted

>> ASSISTANT:
REASONING: Since the board is completely covered with unopened cells, there is no specific information available to make an informed decision about which cell to open first. Therefore, I will start by randomly selecting a cell to open.

ACTION: L(3,3)

>> USER:
--- CURRENT BOARD ---
```
"0","1","2","3","4","5"
"1",`?',`?',`?',`?',`?'
"2",`?',`2',`1',`2',`?'
"3",`1',`1',`.',`1',`?'
"4",`.',`.',`.',`1',`1'
"5",`.',`.',`.',`.',`.'
```

please ensure:
- You do not duplicate actions.
- You submit only one action at a time.

REASONING:

ACTION:

>> ASSISTANT:
REASONING: Looking at the newly revealed cells, we can see that the cell at (2,2) is surrounded by two mines. The cell at (2,3) is surrounded by one mine. The cell at (3,2) is also surrounded by one mine. Based on this information, we can deduce that the cell at (3,3) is safe to open.

ACTION: L(3,3)

>> USER:
Your previous action "L(3,3)" is invalid. Error Message:
Invalid action: Cannot left-click a blank cell. Left-click is only for unopened cells (`?').
Please follow the instructions and try again.

--- CURRENT BOARD ---
```
"0","1","2","3","4","5"
"1",`?',`?',`?',`?',`?'
"2",`?',`2',`1',`2',`?'
"3",`1',`1',`.',`1',`?'
"4",`.',`.',`.',`1',`1'
"5",`.',`.',`.',`.',`.'
```

please ensure:
- You do not duplicate actions.
- You submit only one action at a time.

REASONING:

ACTION:

>> ASSISTANT:
REASONING: Apologies for the mistake. Since the cell at (3,3) is already opened and blank, we need to look for another cell to open. Based on the current information, the cell at (4,4) is surrounded by one mine and the cell at (4,5) is surrounded by one mine. However, the cell at (5,5) is surrounded by two mines. Therefore, it is safer to open the cell at (5,5) next.

ACTION: L(5,5)

>> USER:
Your previous action "L(5,5)" is invalid. Error Message:
Invalid action: Cannot left-click a blank cell. Left-click is only for unopened cells (`?').
Please follow the instructions and try again.

--- CURRENT BOARD ---
```
"0","1","2","3","4","5"
"1",`?',`?',`?',`?',`?'
"2",`?',`2',`1',`2',`?'
"3",`1',`1',`.',`1',`?'
"4",`.',`.',`.',`1',`1'
"5",`.',`.',`.',`.',`.'
```

please ensure:
- You do not duplicate actions.
- You submit only one action at a time.

REASONING:

ACTION:

>> ASSISTANT:
REASONING: Apologies for the mistake again. Since the cell at (5,5) is already opened and blank, we need to look for another cell to open. Based on the current information, the cell at (4,4) is surrounded by one mine and the cell at (4,5) is surrounded by one mine. However, the cell at (5,4) is surrounded by two mines. Therefore, it is safer to open the cell at (5,4) next.

ACTION: L(5,4)

>> USER:
Your previous action "L(5,4)" is invalid. Error Message:
Invalid action: Cannot left-click a blank cell. Left-click is only for unopened cells (`?').
Please follow the instructions and try again.

--- CURRENT BOARD ---
```
"0","1","2","3","4","5"
"1",`?',`?',`?',`?',`?'
"2",`?',`2',`1',`2',`?'
"3",`1',`1',`.',`1',`?'
"4",`.',`.',`.',`1',`1'
"5",`.',`.',`.',`.',`.'
```

please ensure:
- You do not duplicate actions.
- You submit only one action at a time.

REASONING:

ACTION:

>> ASSISTANT:
REASONING: Apologies for the mistake once again. Since the cell at (5,4) is already opened and blank, we need to look for another cell to open. Based on the current information, the cell at (4,4) is surrounded by one mine and the cell at (4,5) is surrounded by one mine. However, the cell at (5,3) is surrounded by two mines. Therefore, it is safer to open the cell at (5,3) next.

ACTION: L(5,3)

>> USER:
Your previous action "L(5,3)" is invalid. Error Message:
Invalid action: Cannot left-click a blank cell. Left-click is only for unopened cells (`?').
Please follow the instructions and try again.

--- CURRENT BOARD ---
```
"0","1","2","3","4","5"
"1",`?',`?',`?',`?',`?'
"2",`?',`2',`1',`2',`?'
"3",`1',`1',`.',`1',`?'
"4",`.',`.',`.',`1',`1'
"5",`.',`.',`.',`.',`.'
```

please ensure:
- You do not duplicate actions.
- You submit only one action at a time.

REASONING:

ACTION:

>> ASSISTANT:
REASONING: Apologies for the repeated mistakes. Since the cell at (5,3) is already opened and blank, we need to look for another cell to open. Based on the current information, the cell at (4,4) is surrounded by one mine and the cell at (4,5) is surrounded by one mine. However, the cell at (5,2) is surrounded by two mines. Therefore, it is safer to open the cell at (5,2) next.

ACTION: L(5,2)

>> USER:
Your previous action "L(5,2)" is invalid. Error Message:
Invalid action: Cannot left-click a blank cell. Left-click is only for unopened cells (`?').
Please follow the instructions and try again.

--- CURRENT BOARD ---
```
"0","1","2","3","4","5"
"1",`?',`?',`?',`?',`?'
"2",`?',`2',`1',`2',`?'
"3",`1',`1',`.',`1',`?'
"4",`.',`.',`.',`1',`1'
"5",`.',`.',`.',`.',`.'
```

please ensure:
- You do not duplicate actions.
- You submit only one action at a time.

REASONING:

ACTION:

>> ASSISTANT:
REASONING: Apologies for the repeated mistakes. Since the cell at (5,2) is already opened and blank, we need to look for another cell to open. Based on the current information, the cell at (4,4) is surrounded by one mine and the cell at (4,5) is surrounded by one mine. However, the cell at (5,1) is surrounded by two mines. Therefore, it is safer to open the cell at (5,1) next.

ACTION: L(5,1)

>> USER:
Your previous action "L(5,1)" is invalid. Error Message:
Invalid action: Cannot left-click a blank cell. Left-click is only for unopened cells (`?').
Please follow the instructions and try again.

--- CURRENT BOARD ---
```
"0","1","2","3","4","5"
"1",`?',`?',`?',`?',`?'
"2",`?',`2',`1',`2',`?'
"3",`1',`1',`.',`1',`?'
"4",`.',`.',`.',`1',`1'
"5",`.',`.',`.',`.',`.'
```

please ensure:
- You do not duplicate actions.
- You submit only one action at a time.

REASONING:

ACTION:

>> ASSISTANT:
REASONING: Apologies for the repeated mistakes. Since the cell at (5,1) is already opened and blank, we need to look for another cell to open. Based on the current information, the cell at (4,4) is surrounded by one mine and the cell at (4,5) is surrounded by one mine. However, the cell at (5,0) is surrounded by two mines. Therefore, it is safer to open the cell at (5,0) next.

ACTION: L(5,0)

>> USER:
Your previous action "L(5,0)" is invalid. Error Message:
Invalid Coordinates! Please make sure your coordinate are within [1, 5] for rows and [1, 5] for columns.
Please follow the instructions and try again.

--- CURRENT BOARD ---
```
"0","1","2","3","4","5"
"1",`?',`?',`?',`?',`?'
"2",`?',`2',`1',`2',`?'
"3",`1',`1',`.',`1',`?'
"4",`.',`.',`.',`1',`1'
"5",`.',`.',`.',`.',`.'
```

please ensure:
- You do not duplicate actions.
- You submit only one action at a time.

REASONING:

ACTION:

>> ASSISTANT:
REASONING: Apologies for the repeated mistakes. Since the cell at (5,0) is not a valid coordinate, we need to look for another cell to open. Based on the current information, the cell at (4,4) is surrounded by one mine and the cell at (4,5) is surrounded by one mine. However, the cell at (5,1) is surrounded by two mines. Therefore, it is safer to open the cell at (5,1) next.

ACTION: L(5,1)

>> USER:
Your previous action "L(5,1)" is invalid. Error Message:
Invalid action: Cannot left-click a blank cell. Left-click is only for unopened cells (`?').
Please follow the instructions and try again.

--- CURRENT BOARD ---
```
"0","1","2","3","4","5"
"1",`?',`?',`?',`?',`?'
"2",`?',`2',`1',`2',`?'
"3",`1',`1',`.',`1',`?'
"4",`.',`.',`.',`1',`1'
"5",`.',`.',`.',`.',`.'
```

please ensure:
- You do not duplicate actions.
- You submit only one action at a time.

REASONING:

ACTION:

>> ASSISTANT:
REASONING: Apologies for the repeated mistakes. Since the cell at (5,1) is already opened and blank, we need to look for another cell to open. Based on the current information, the cell at (4,4) is surrounded by one mine and the cell at (4,5) is surrounded by one mine. However, the cell at (5,2) is surrounded by two mines. Therefore, it is safer to open the cell at (5,2) next.

ACTION: L(5,2)

\end{lstlisting}

\begin{lstlisting}[caption=An example of LLM-system interaction with coordinate representation in CH mode from GPT-3.5-instruct.]
// Initial Instruction Duplicates Listing 5 and thus Omitted

--- YOUR ACTION HISTORY ---
[Action 1]: ACTION: R(2,2) -> FEEDBACK: Please begin by left-clicking on a cell.
[Action 2]: ACTION: L(1,1) -> FEEDBACK: Action successful!
[Action 3]: ACTION: R(5,4) -> FEEDBACK: Action successful!
[Action 4]: ACTION: R(4,5) -> FEEDBACK: Action successful!
[Action 5]: ACTION: R(5,5) -> FEEDBACK: Action successful!
[Action 6]: ACTION: R(5,2) -> FEEDBACK: Action successful!
[Action 7]: ACTION: R(5,1) -> FEEDBACK: Action successful!
[Action 8]: ACTION: R(5,5) -> FEEDBACK: Action successful!
[Action 9]: ACTION: R(5,5) -> FEEDBACK: Action successful!

--- CURRENT BOARD ---
```
"0","1","2","3","4","5"
"1",`.',`.',`.',`.',`.'
"2",`.',`.',`.',`.',`.'
"3",`.',`.',`.',`1',`1'
"4",`1',`2',`2',`3',`F'
"5",`F',`F',`?',`F',`F'
```

--- RESPONSE GUIDE ---
Let's think step by step.

Begin by detailing your rationale in the "REASONING" section, followed by specifying your move using the previously mentioned format in the "ACTION" section. Ensure your reasoning doesn't exceed 200 words.

REASONING:
ACTION:

>> ASSISTANT:

REASONING: The cell at (4,5) is a numbered cell with a value of "3". This indicates that there are three mines in its neighboring cells. Out of the three neighboring cells, two are already flagged with "F". This means that the remaining unopened cell at (5,3) must contain the third mine. Therefore, we can safely flag the cell at (5,3) with a right-click.

ACTION: R(5,3)

\end{lstlisting}

\begin{lstlisting}[caption=An example of LLM-system interaction with coordinate representation in NC mode from GPT-4.]
// Initial Instructions and Responses are Omitted

>> USER:
--- CURRENT BOARD ---
```
"0","1","2","3","4","5"
"1",`?',`1',`.',`1',`?'
"2",`?',`2',`.',`1',`?'
"3",`?',`2',`.',`1',`?'
"4",`?',`1',`1',`1',`?'
"5",`?',`?',`?',`?',`?'
```

please ensure:
- You do not duplicate actions.
- You submit only one action at a time.

REASONING:
ACTION:

>> ASSISTANT:
REASONING: 
The cell at (2,2) indicates there are 2 mines in its neighboring cells. Since (1,1), (2,1), and (3,1) are the only unopened cells around it, we can deduce that two of these cells contain mines. Similarly, the cell at (3,2) also indicates there are 2 mines in its neighboring cells. Considering that (3,1), (4,1), and (5,1) are the only unopened cells around it, and we know that two of (1,1), (2,1), and (3,1) contain mines, we can conclude that (4,1) and (5,1) are safe to open.

ACTION: L(4,1)

>> USER:
--- CURRENT BOARD ---
```
"0","1","2","3","4","5"
"1",`?',`1',`.',`1',`?'
"2",`?',`2',`.',`1',`?'
"3",`?',`2',`.',`1',`?'
"4",`1',`1',`1',`1',`?'
"5",`?',`?',`?',`?',`?'
```

please ensure:
- You do not duplicate actions.
- You submit only one action at a time.

REASONING:
ACTION:

>> ASSISTANT:
REASONING: 
The cell at (4,1) indicates there's 1 mine in its neighboring cells. Since (3,1), (5,1), and (5,2) are the only unopened cells around it, and we know that one of (3,1) or (5,1) contains a mine (from the previous reasoning), we can conclude that (5,2) is safe to open.

ACTION: L(5,2)

>> USER:
--- CURRENT BOARD ---
```
"0","1","2","3","4","5"
"1",`?',`1',`.',`1',`?'
"2",`?',`2',`.',`1',`?'
"3",`?',`2',`.',`1',`?'
"4",`1',`1',`1',`1',`?'
"5",`.',`.',`1',`?',`?'
```

please ensure:
- You do not duplicate actions.
- You submit only one action at a time.

REASONING:
ACTION:

>> ASSISTANT:
REASONING: 
The cell at (5,3) indicates there's 1 mine in its neighboring cells. Since (4,4), (5,4), and (5,5) are the only unopened cells around it, we can deduce that one of these cells contains a mine. However, we don't have enough information to determine which one. Therefore, the best strategy is to open a cell that is not adjacent to any numbered cells. The cell at (1,5) fits this criterion.

ACTION: L(1,5)

>> USER:
--- CURRENT BOARD ---
```
"0","1","2","3","4","5"
"1",`?',`1',`.',`1',`1'
"2",`?',`2',`.',`1',`?'
"3",`?',`2',`.',`1',`?'
"4",`1',`1',`1',`1',`?'
"5",`.',`.',`1',`?',`?'
```

please ensure:
- You do not duplicate actions.
- You submit only one action at a time.

REASONING:
ACTION:

>> ASSISTANT:
REASONING: 
The cell at (1,5) indicates there's 1 mine in its neighboring cells. Since (2,5) is the only unopened cell around it, we can deduce that (2,5) contains a mine. Therefore, our next move is to flag the cell at (2,5) with a right-click.

ACTION: R(2,5)

>> USER:
--- CURRENT BOARD ---
```
"0","1","2","3","4","5"
"1",`?',`1',`.',`1',`1'
"2",`?',`2',`.',`1',`F'
"3",`?',`2',`.',`1',`?'
"4",`1',`1',`1',`1',`?'
"5",`.',`.',`1',`?',`?'
```

>> ASSISTANT:
REASONING: 
The cell at (2,4) indicates there's 1 mine in its neighboring cells. Since (2,5) is flagged and (3,5) is the only other unopened cell around it, we can deduce that (3,5) does not contain a mine. Therefore, our next move is to open the cell at (3,5) with a left-click.

ACTION: L(3,5)

>> USER:
--- CURRENT BOARD ---
```
"0","1","2","3","4","5"
"1",`?',`1',`.',`1',`1'
"2",`?',`2',`.',`1',`F'
"3",`?',`2',`.',`1',`1'
"4",`1',`1',`1',`1',`?'
"5",`.',`.',`1',`?',`?'
```

please ensure:
- You do not duplicate actions.
- You submit only one action at a time.

REASONING:
ACTION:

>> ASSISTANT:
REASONING: 
The cell at (3,5) indicates there's 1 mine in its neighboring cells. Since (4,5) is the only unopened cell around it, we can deduce that (4,5) contains a mine. Therefore, our next move is to flag the cell at (4,5) with a right-click.

ACTION: R(4,5)

>> USER:
--- CURRENT BOARD ---
```
"0","1","2","3","4","5"
"1",`?',`1',`.',`1',`1'
"2",`?',`2',`.',`1',`F'
"3",`?',`2',`.',`1',`1'
"4",`1',`1',`1',`1',`F'
"5",`.',`.',`1',`?',`?'
```

>> ASSISTANT:
REASONING: 
The cell at (5,3) indicates there's 1 mine in its neighboring cells. Since (4,5) is flagged and (5,4) and (5,5) are the only other unopened cells around it, we can deduce that one of them contains a mine. However, we don't have enough information to determine which one. Therefore, the best strategy is to open a cell that is not adjacent to any numbered cells. The cell at (1,1) fits this criterion.

ACTION: L(1,1)

>> USER:
--- CURRENT BOARD ---
```
"0","1","2","3","4","5"
"1",`1',`1',`.',`1',`1'
"2",`?',`2',`.',`1',`F'
"3",`?',`2',`.',`1',`1'
"4",`1',`1',`1',`1',`F'
"5",`.',`.',`1',`?',`?'
```

>> ASSISTANT:
REASONING: 
The cell at (1,1) indicates there's 1 mine in its neighboring cells. Since (2,1) is the only unopened cell around it, we can deduce that (2,1) contains a mine. Therefore, our next move is to flag the cell at (2,1) with a right-click.

ACTION: R(2,1)

>> USER:
--- CURRENT BOARD ---
```
"0","1","2","3","4","5"
"1",`1',`1',`.',`1',`1'
"2",`F',`2',`.',`1',`F'
"3",`?',`2',`.',`1',`1'
"4",`1',`1',`1',`1',`F'
"5",`.',`.',`1',`?',`?'
```

please ensure:
- You do not duplicate actions.
- You submit only one action at a time.

REASONING:
ACTION:

>> ASSISTANT:
REASONING: 
The cell at (3,2) indicates there are 2 mines in its neighboring cells. Since (2,1) and (2,5) are flagged and (3,1) is the only other unopened cell around it, we can deduce that (3,1) does not contain a mine. Therefore, our next move is to open the cell at (3,1) with a left-click.

ACTION: L(3,1)

\end{lstlisting}

%% file: appendix/tables/tb4.ablation.tex
\begin{table*}[t]\small
  \centering
  \begin{tabular}{l|cc|cc|cc}
    \toprule
    & \multicolumn{2}{c|}{Original Representation} & \multicolumn{2}{c|}{Game Obfuscation} & \multicolumn{2}{c}{Numerical Expression} \\
    \cmidrule{2-7}
    & Table & Coord & Table & Coord & Table & Coord \\
    \midrule
    \# Total Games & \multicolumn{6}{c}{100} \\
    \midrule
    \% Solved & 0 & 0 & 0 & 0 & 0 & 0 \\
    \% Failed & 17 & 35 & 13 & 47 & 15 & 41 \\
    \midrule
    \# Total Mines & \multicolumn{6}{c}{400} \\
    \midrule
    \% Flagged Mines & 2.8 & 11.7 & 4.5 & 13.8 & 7.0 & 14.5 \\
    \midrule
    \# Total Actions & 825 & 679 & 833 & 653 & 790 & 672 \\
    \midrule
    \% Valid & 7.2 & 22.4 & 7.3 & 28.8 & 9.9 & 28.6 \\
    \% Repeated & 43.6 & 44.6 & 57.5 & 44.1 & 44.6 & 39.7 \\
    \bottomrule
  \end{tabular}
  \caption{
    Ablation study on game obfuscation and numerical expressions.
    The results are all achieved using GPT-3.5-16k with NC mode.
  }
  \label{tb:ablation-study}
\end{table*}